\pdfoutput=1
\newcommand\coffee{\raisebox{-2pt}{\includegraphics[width=1em]{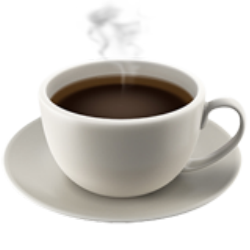}}}
\documentclass[11pt]{article}

\usepackage[]{acl}

\usepackage{colortbl}
\usepackage{xcolor}
\usepackage{makecell}

\usepackage{times}
\usepackage{latexsym}

\usepackage[T1]{fontenc}

\usepackage[utf8]{inputenc}

\usepackage{microtype}

\usepackage{inconsolata}

\usepackage{graphicx}
\usepackage{algorithm}
\usepackage[noend]{algpseudocode}
\usepackage{booktabs}
\usepackage{algpseudocode}
\usepackage{subcaption}
\usepackage{tabularx}
\usepackage{multirow}
\usepackage{amsmath}

\newcommand{\ie}{\textit{i.e.}}
\newcommand{\eg}{\textit{e.g.}}

\title{Commonsense-augmented Memory Construction and Management in Long-term Conversations via Context-aware Persona Refinement
}

\author{
    Hana Kim$^{1}$~~~
    Kai Tzu-iunn Ong$^{2}$~~~
    Seoyeon Kim$^{2}$~~~
    Dongha Lee$^{2}$~~~
    Jinyoung Yeo$^{2}$\\
    Department of Computer Science$^{1}$, Artificial Intelligence$^{2}$, Yonsei University\\
    \texttt{\{hana.kim,ktio89,emseoyk,donalee,jinyeo\}@yonsei.ac.kr}\\
}

\begin{document}
\maketitle
\begin{abstract}
Memorizing and utilizing speakers' personas is a common practice for response generation in long-term conversations. Yet, human-authored datasets often provide uninformative persona sentences that hinder response quality. This paper presents a novel framework that leverages commonsense-based persona expansion to address such issues in long-term conversation.
While prior work focuses on not producing personas that contradict others, we focus on transforming contradictory personas into sentences that contain rich speaker information, by refining them based on their contextual backgrounds with designed strategies. As the pioneer of persona expansion in multi-session settings, our framework facilitates better response generation via human-like persona refinement. The supplementary video of our work is available at \url{https://caffeine-15bbf.web.app/}. 

\end{abstract}

\section{Introduction}

Memorizing participants' personal information and conversing accordingly is important for dialogue systems to maintain long-term intimacy with users~\citep{adiwardana2020towards}.
For that, studies have proposed datasets of long-term conversations, which require dialogue systems to memorize and utilize speakers' personas from past dialogue sessions to generate proper responses~\citep{xu2021beyond, bae2022keep}.
Regardless, human-authored personas can be generic and over-simplified, limiting the generation of diverse and engaging responses.

\begin{figure}[t]
\centering
    \includegraphics[width=1\columnwidth]{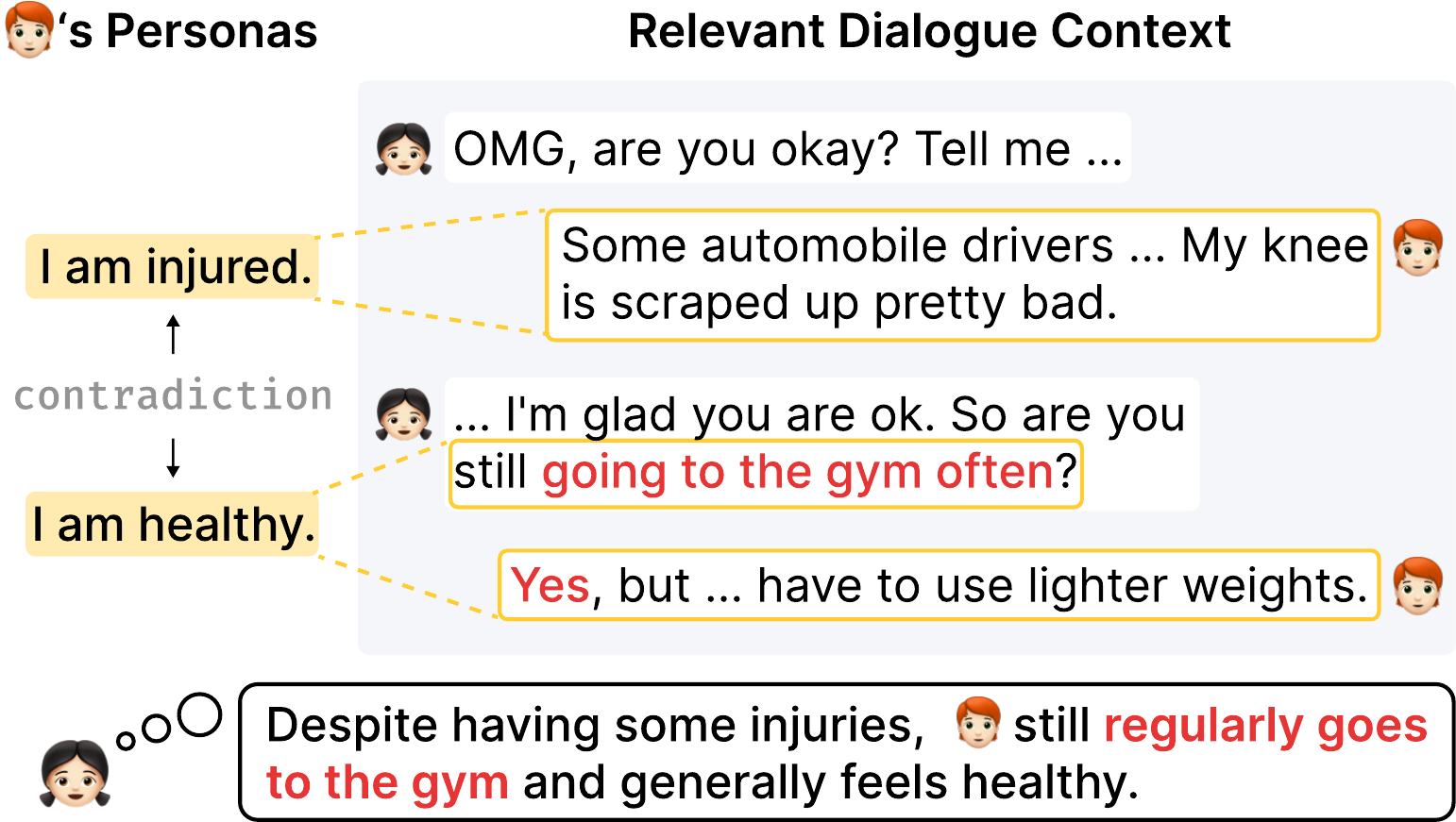}
    \caption{Contradictory personas can co-exist and provide rich speaker information for the conversation when their contexts are considered (an empirical example).}
    \label{fig:example_first_page} 
\end{figure}

Intuitively, this can be addressed by expanding existing personas with commonsense expansion~\cite{majumder2020like}. However, such a naive remedy can raise contradiction between personas (\eg, \textit{``I am lazy''} and \textit{``I clean my room every day''}), especially as sessions accumulate (Figure~\ref{fig:both-figures}), hindering consistent response generation. While we can simply get rid of contradictory personas utilizing external modules such as models for natural language inference (NLI), it yields sub-optimal results (Section~\ref{ssec:results}). Also, avoiding contradictory personas~\citep{ bae2022keep, kim2023persona} does not align with human personality traits. Since human personality is context-dependent~\citep{van2005context}, we naturally exhibit different personalities and behaviors in different contexts, allowing personas with contradictory interpretations to coexist as one's personas, as shown in Figure~\ref{fig:example_first_page}.

\begin{figure*}[th]
    \centering
    \includegraphics[width=1\textwidth]{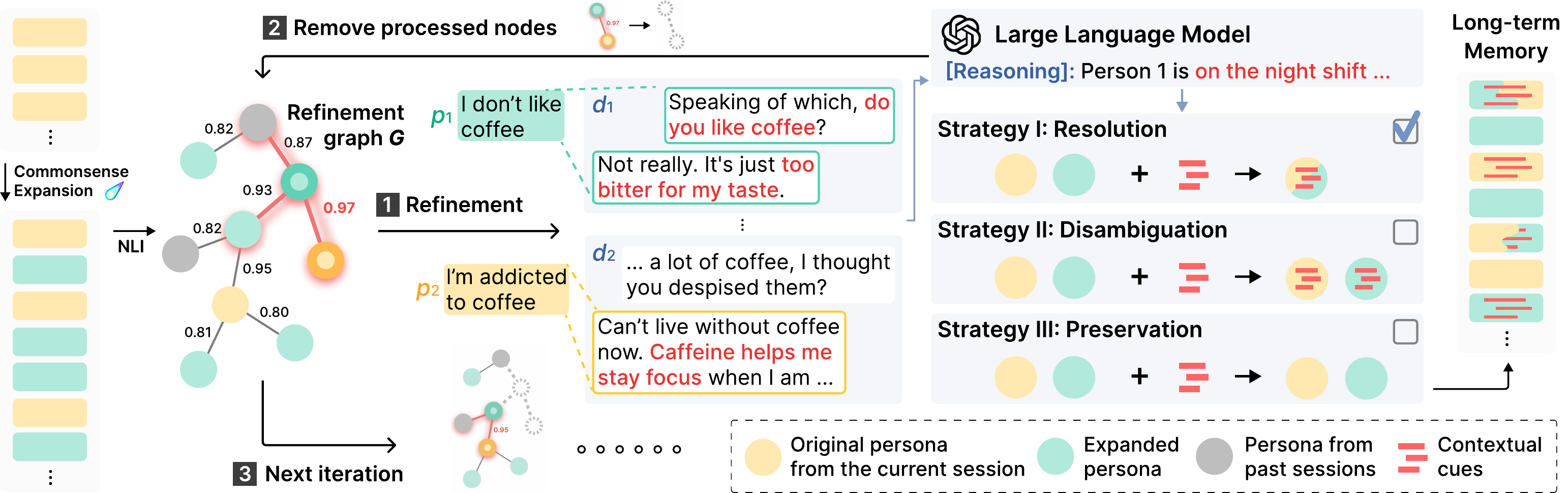}
    \caption{At the end of each dialogue session, \textsc{Caffeine} refines contradictory personas within/across the session(s) and saves the refined version to the dialogue model's memory for response generation in the next session.}
    \label{fig:overview} 
\end{figure*}

Motivated by these, in this paper, we tackle such bottleneck of persona expansion in long-term conversations. Specifically, we focus on transforming contradictory personas into sentences that contain richer speaker information. To this end, we present \coffee{} \textbf{\textsc{Caffeine}}, a \textbf{C}ontext-\textbf{A}ware re\textbf{F}inement \textbf{F}ramework for contradictory p\textbf{E}rsonas \textbf{IN} long-t\textbf{E}rm conversations. \textsc{Caffeine} leverages large language models (LLMs) to iteratively refine the contradictory personas within/across the session(s) based on their contextual background with designed strategies.
Our contributions are two-fold: 
(i) To the best of our knowledge, we are the first to explore commonsense-based persona expansion in multi-session settings; 
(ii) \textsc{Caffeine} enables better response generation in long-term conversations in both automatic and human evaluations. Also, it refines contradictory personas in a human-like manner, eliciting persona sentences that are superior in various criteria while being cost- and time-efficient.

\section{Approach}
Long-term conversations involve multiple dialogue sessions. At the end of each session, we perform:

\subsection{Commonsense-based Persona Expansion}

Following~\citet{majumder2020like}, we perform commonsense expansion on personas derived from the conversation using COMET~\citep{hwang2021comet}. COMET generates commonsense knowledge based on cause-effect relation types (\eg, \textsc{xNeed} and \textsc{xWant}). 
For example, \textit{``I drink coffee''} $\rightarrow$ \textit{``I want to stay awake''}. Implementation details on COMET expansion are in Appendix~\ref{ssec:app_comet}.

\subsection{\coffee{} \textsc{Caffeine}}
We present the overview of \textsc{Caffeine} in Figure~\ref{fig:overview}.

\subsubsection{Preparation: Graph Construction for Iterative Persona Refinement}

After expansion, we identify contradictory personas by computing the probability of contradiction $\delta$  between all personas with an external NLI model. 
To refine contradictory personas cost- and time-efficiently, we adopt iterative refinement with a graph structure: Contradictory pairs with $\delta$ larger than a threshold $\mu$ are added as nodes $V$ (edges $E = (\delta_1, \delta_2, \cdots, \delta_{|E|})$ ) to the refinement graph $G$.\footnote{We empirically set $\mu$ as 0.8.} Then, we locate the node (persona) $p_1$ with the largest $\Sigma \delta$ within its neighborhood. We select $p_1$ and the adjacent node $p_2$ with the highest $\delta$ with $p_1$ for the first refinement iteration (Algorithm~\ref{alg:graph_algo}).

\subsubsection{Context-aware Persona Refinement}
\begin{figure}[h]
    \centering
    \includegraphics[width=1\columnwidth]{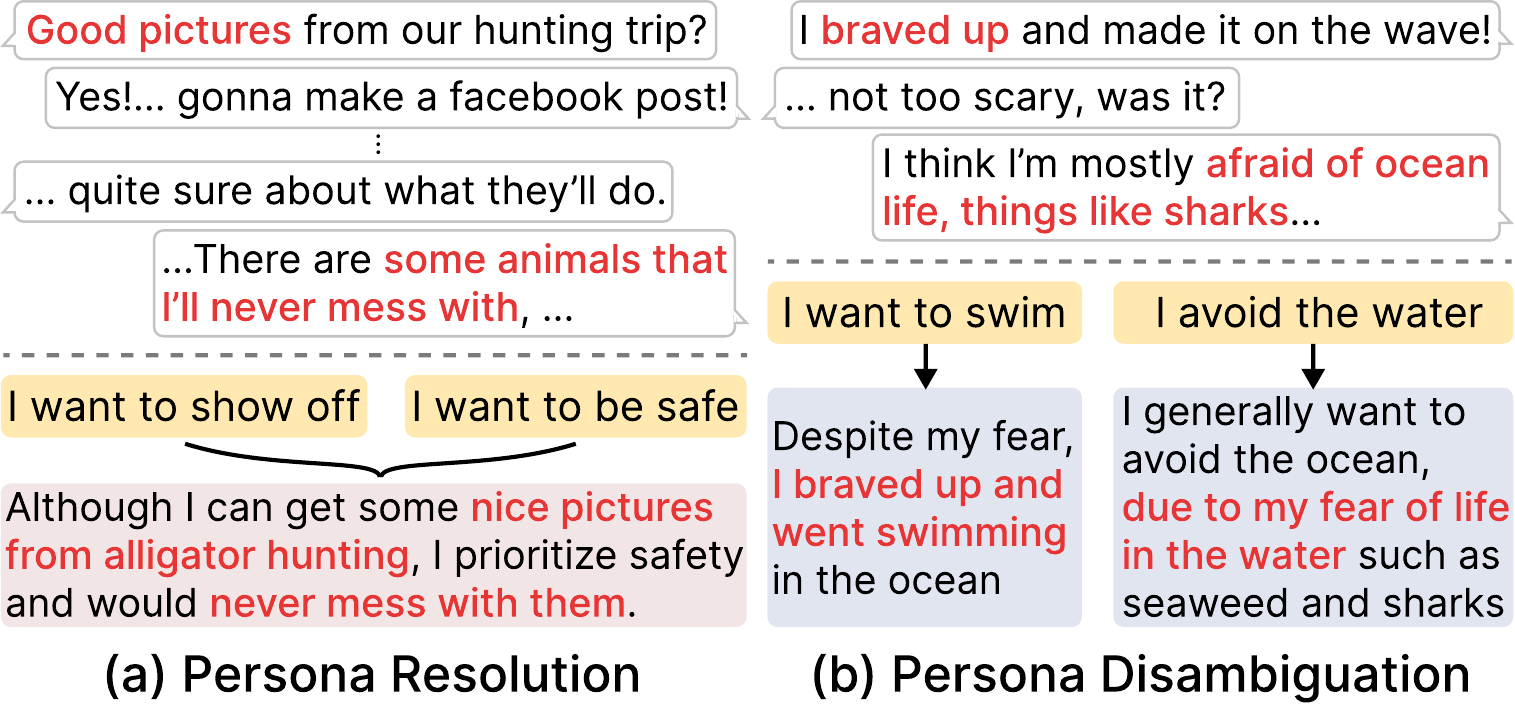}
    \caption{Empirical demonstration of our strategies. Top: relevant contexts; Mid: contradictory personas; Bottom: refined persona(s).}
    \label{fig:method_example}
\end{figure}

As shown in Figure~\ref{fig:example_first_page}, personas causing contradiction can be logically acceptable and beneficial for conversations if contextual cues from their origin context are appended via commonsense reasoning. For that, we propose the following refinement strategies for the LLM to choose from:

\paragraph{Strategy I: Resolution.}
Inspired by entity resolution~\citep{benjelloun2009swoosh}, persona resolution resolves the contradiction between personas by seamlessly \textbf{merging} them into one informative sentence based on the contextual background from where they are derived (Figure~\ref{fig:method_example} (a), Figure~\ref{fig:example_first_page} is also an example of persona resolution).

\paragraph{Strategy II: Disambiguation.}
Contradiction between two statements can stem from the lack of contexts, known as pragmatic ambiguity~\citep{macagno2018types}.
Drawn from entity disambiguation~\citep{dredze2010entity}, persona disambiguation \textbf{specifies} each persona with relevant information from their contextual backgrounds (Figure~\ref{fig:method_example} (b)).

\begin{table*}[ht]
\small
\centering
\begin{tabular}{lcccccccccccc}
\toprule
 &
  \multicolumn{3}{c}{\textbf{Session 2}} &
  \multicolumn{3}{c}{\textbf{Session 3}} &
  \multicolumn{3}{c}{\textbf{Session 4}} &
  \multicolumn{3}{c}{\textbf{Session 5}} \\ \cmidrule(lr){2-4} \cmidrule(lr){5-7} \cmidrule(lr){8-10} \cmidrule(l){11-13} 
\textbf{Settings} &
  B-1 &
  R-1 &
  R-L &
  B-1 &
  R-1 &
  R-L &
  B-1 &
  R-1 &
  R-L &
  B-1 &
  R-1 &
  R-L \\ 
\toprule

No Memory & 20.75 & 19.38 & 15.16 & 20.42 & 19.53 & 15.09 & 19.88 & 19.56 & 14.98 & 19.87 & 20.16 & 15.33 \\ 
\midrule
GOLD & \underline{21.19} & 19.86 & \textbf{15.50} & 21.24 & 20.16 & 15.47 & 20.57 & 19.94 & 15.16 & 20.49 & 20.53 & 15.55 \\ 
\hspace{0.01cm}+ NLI-remove & 20.81 & 19.98 & 15.26 & 21.04 & 20.28 & 15.52 & 21.33 & 20.69 & 15.91 & 21.43 & 20.75 & 15.95 \\ 
\hspace{0.01cm}+ NLI-recent & 20.87 & \underline{20.09} & 15.39 & 21.14 & 20.52 & \underline{15.71} & 21.46 & 20.79 & 15.97 & 21.60 & 20.97 & 16.11 \\ 
\cellcolor{gray!13}\hspace{0.01cm}\textbf{+ \textsc{Caffeine}} & \cellcolor{gray!13}20.93 & \cellcolor{gray!13}\textbf{20.18} & \cellcolor{gray!13}\underline{15.47} & \cellcolor{gray!13}\underline{21.41} & \cellcolor{gray!13}\underline{20.72} & \cellcolor{gray!13}\textbf{15.86} & \cellcolor{gray!13}\underline{21.67} & \cellcolor{gray!13}\underline{21.00} & \cellcolor{gray!13}\underline{16.15} & \cellcolor{gray!13}\underline{21.92} & \cellcolor{gray!13}\underline{21.23} & \cellcolor{gray!13}\underline{16.31} \\ \midrule
COMET-EXP & \textbf{21.23} & 19.82 & 15.44 & 20.95 & 19.90 & 15.38 & 20.33 & 20.02 & 15.18 & 20.00 & 20.27 & 15.37 \\ 
\hspace{0.01cm}+ NLI-remove & 20.72 & 19.96 & 15.27 & 21.12 & 20.40 & 15.56 & 21.66 & 20.77 & 15.88 & 21.77 & 20.91 & 16.01 \\ 
\hspace{0.01cm}+ NLI-recent & 20.73 & 20.00 & 15.33 & 21.16 & 20.40 & 15.64 & 21.57 & 20.77 & 15.89 & 21.78 & 20.99 & 16.09 \\ 
\cellcolor{gray!13}\hspace{0.01cm}\textbf{+ \textsc{Caffeine}} & \cellcolor{gray!13}20.97 & \cellcolor{gray!13}20.06 & \cellcolor{gray!13}15.32 & \cellcolor{gray!13}\textbf{21.63} & \cellcolor{gray!13}\textbf{20.73} & \cellcolor{gray!13}\textbf{15.86} & \cellcolor{gray!13}\textbf{21.97} & \cellcolor{gray!13}\textbf{21.10} & \cellcolor{gray!13}\textbf{16.18} & \cellcolor{gray!13}\textbf{22.26} & \cellcolor{gray!13}\textbf{21.32} & \cellcolor{gray!13}\textbf{16.37} \\ 

\bottomrule
\end{tabular}
\caption{Performance in response generation. \textbf{Bold} and \underline{underline} show the best and second-highest in each column.}
\label{tab:main_results}
\end{table*}
\paragraph{Strategy III: Preservation.}
Due to the limitation of NLI models, personas predicted as contradictory may be consistent and may not require refinement. Thus, we allow the LLM to preserve personas as they are when their contexts suggest so. 

In practice, with contradictory personas $\mathcal{P} = (p_1, p_2)$ and relevant dialogue contexts $\mathcal{D} = (d_1, d_2)$ from where $\mathcal{P}$ are derived, we prompt the LLM to choose one out of the three strategies $S$ with rationale and generate the refinement $\mathcal{R}$:
\begin{align}
\label{eq:extract_cot_2}
    \mathcal{S}^* = \underset{\mathcal{S}}{\text{argmax}}\,  P_{\text{LLM}}(\mathcal{S}|\mathcal{P}, \mathcal{D}) \\
    \Rightarrow \, \mathcal{R}^* = \underset{\mathcal{R}}{\text{argmax}}\,
    P_{\text{LLM}}(\mathcal{R}|\mathcal{P}, \mathcal{D}, \mathcal{S}^*)
\end{align}
where $\Rightarrow$ denotes a sequential generation of tokens. $D$ consists of $w$ consecutive utterances.\footnote{In our experiments, $w$ differs depending on the persona annotation in the applied dataset. See Appendix~\ref{ssec:dialoguefragment}.} When $p$ is a persona generated by COMET, we use $D$ of its original persona and concatenate the original persona with $D$.
After refinement, we save $\mathcal{R}^*$ to long-term memory $\mathcal{M}$ and remove $\mathcal{P} $ from the graph $G$, and start the next iteration (Algorithm~\ref{alg:graph_algo}).

\begin{algorithm}[h]
\small
\caption{Iterative Graph Refinement}
\begin{algorithmic}[1]
\Require Refinement graph \( G(V, E) \)
\Ensure The dialogue model's long-term memory $\mathcal{M}$ 
\State \( \mathcal{M} \leftarrow \) $\mathcal{M} \setminus V$
\While{\( G \neq \emptyset \)}
    \State Select \( p_1 \) in \( V \) with the highest \( \Sigma \delta \)
    \State Select \( p_2 \), a neighbor of \( p_1 \) with the highest \( \delta \)
    \State \( (\mathcal{S}^*, \mathcal{R}^*) \leftarrow \mathrm{Refine}(p_1, p_2)\) 
    \State \( \mathcal{M} \leftarrow \) $\mathcal{M}\cup\mathcal{R}^*$
    \State Remove \( p_1, p_2 \) from \( G \)
    \State Remove isolated nodes from \( G \)
\EndWhile
\State \Return $\mathcal{M}$
\end{algorithmic}
\label{alg:graph_algo}
\end{algorithm}

\section{Experiments}
\label{sec:experiments}

\subsection{Experimental Settings}
\paragraph{Dataset.} We use Multi-Session Chat (MSC)~\citep{xu2021beyond} to conduct experiments. MSC takes the dialogues from Persona-Chat~\citep{zhang2018personalizing} and extends their follow-up conversations throughout several sessions. Each session comes with speakers' personas authored by humans.

\paragraph{Models and baselines.}
In this work, we use ChatGPT~\cite{openai2023chatgpt} for \textsc{Caffeine} and response generation (RG),\footnote{Prompts for RG and \textsc{Caffeine} are in Appendix~\ref{ssec:prompt}.} and Contriever~\cite{izacard2021unsupervised} to retrieve top-$k$ relevant personas from long-term memory.\footnote{We set $k =$ 20. Results with other $k$ are in Appendix~\ref{ssec:topk}.} As for the NLI model, we use RoBERTa~\cite{liu2019roberta} fine-tuned on the MNLI dataset~\cite{williams2017broad}. To evaluate the effectiveness of \textsc{Caffeine} in RG, we apply it to: (i) COMET-EXP, human-authored personas with COMET expansion; (ii) GOLD, human-authored personas. We include this setting as a contradiction can also exist among un-expanded personas.\footnote{We report the statistics of contradiction in Appendix~\ref{ssec:nli}. As our focus is persona expansion, extracting personas from conversations is out of the scope of this work.} Also, to justify our choice to refine rather than remove, we compare \textsc{Caffeine} with two baselines: NLI-remove and NLI-recent.\footnote{We test with MNLI and DNLI~\cite{welleck-etal-2019-dialogue} and report results using MNLI as it shows better performance. Results with DNLI are in Appendix~\ref{ssec:nli}.} The NLI-remove approach filters out personas that contradict at least one other persona with $\delta \ge 0.8$ via the NLI model. Similarly, the NLI-recent approach also uses the NLI model, but it differs by keeping the most recent persona in contradictory persona pairs and removing the older one~\cite{bae2022keep}, thereby prioritizing updated personas over time.

\subsection{Results and Discussion}
\label{ssec:results}
We present the empirical findings of the following research
questions guiding our experiments:

\noindent \textbf{RQ1}: \textit{Does \textsc{Caffeine} benefit response generation in long-term conversations?}

\noindent \textbf{RQ2}: \textit{Does \textsc{Caffeine} refine personas in a way that aligns with human judgment?}

\noindent \textbf{RQ3}: \textit{Is \textsc{Caffeine} cost- and time-efficient?}

\paragraph{\textsc{Caffeine} improves response generation (RQ1).} 
To evaluate the efficacy of \textsc{Caffeine}, we conduct experiments on response generation (RG) using sessions 2 to 5 of each dialogue from MSC.
Table~\ref{tab:main_results} shows the results of RG in MSC with BLEU-1 (B-1), ROUGE-1 (R-1), and ROUGE-L (R-L)~\cite{papineni2002bleu, lin2004rouge}.
Applying \textsc{Caffeine} yields performance gains, which are more significant as sessions increase. 
Also, \textsc{Caffeine} consistently outperforms NLI-remove and NLI-recent, showing that leveraging contradictory personas elicits a more informative memory for RG than removing them. Compared to NLI-remove, the improved efficacy of NLI-recent is attributed to its focus on the recency of personas. By eliminating outdated personas from contradictory pairs, NLI-recent enhances RG, yielding responses more aligned with the current dialogue context. However, despite the enhancements in NLI-recent performance, \textsc{Caffeine} still exhibits superior performance. Furthermore, the performance brought by \textsc{Caffeine} exhibits a continuously rising trend as the number of previous sessions increases, while baselines yield a flat or downward tendency.
These demonstrate the effectiveness of \textsc{Caffeine} in multi-session conversations. Table~\ref{tab:filter_passed_and_failed} shows the human evaluation results of randomly sampled 50 responses conducted by 3 judges from Amazon Mechanical Turk (Appendix~\ref{sec:AMT}). \textsc{Caffeine} yields responses that are better (\ie, winning) in several criteria. We provide examples of RG in Appendix~\ref{sec:app_examples}.

\begin{table}[t]
\setlength{\tabcolsep}{3pt}
\centering
\small
\begin{tabular}{l|ccc}

\toprule
{\textbf{\textsc{Caffeine}}} vs.
 & \textbf{GOLD} & \textbf{COMET-EXP} & \textbf{NLI-remove}  \\
\midrule
Naturalness  & 73\%$^*$& 71\%$^*$  & 79\%$^*$ \\
Consistency   & 66\%$^*$ & 62\%$^*$   & 67\%$^*$  \\
Specificity  & 55\%~~ & 53\%~~ & 51\%~~ \\ 
Engagingness  & 63\%$^*$ & 64\%$^*$  & 66\%$^*$   \\
Overall  & 62\%$^*$ & 63\%$^*$  & 67\%$^*$    \\
\bottomrule
\end{tabular}
\caption{Comparison of generated responses. We report \textsc{Caffeine}'s \textbf{winning rate}. (*: p-value $< 0.05$)}
\label{tab:filter_passed_and_failed}
\end{table}

\begin{figure}[t]
    \centering
    \includegraphics[width=1\columnwidth]{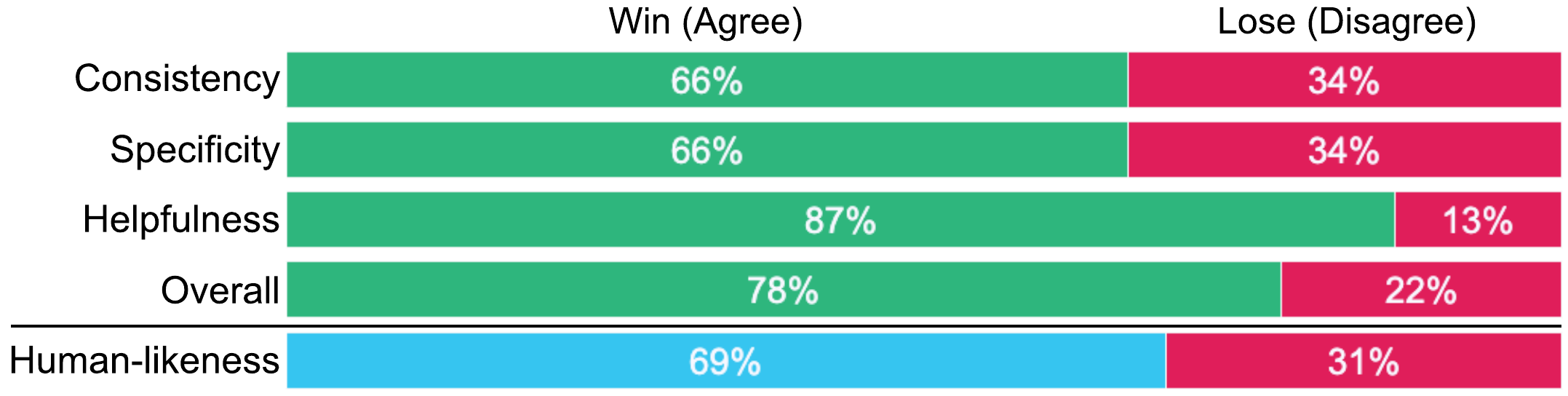}
    \caption{Human evaluation results on (i) refined personas and (ii) the refinement process (p-value $< 0.05$).}
    \label{fig:RG_refine}
\end{figure}

\paragraph{\textsc{Caffeine} elicits personas that align with human preference (RQ2).}
We sample 100 persona pairs refined with ``resolution'' or ``disambiguation'' and ask 3 judges \textit{`` whether they are contradictory before refinement from a human standpoint''}. 89 samples that receive \textit{``yes''} from all judges are used for the evaluation. Judges compare the refined version with its un-refined version and vote if they agree: it is less contradictory (Consistency); it provides more speaker information (Specificity); it is more useful when having a conversation with this person (Helpfulness); it has better quality (Overall); the refinement process is reasonable (Human-likeness).
Figure~\ref{fig:RG_refine} shows that personas refined by \textsc{Caffeine} are greater in all criteria, especially helpfulness. This supports our argument that contradictory personas become sentences with rich speaker information for the conversation if cues from their relevant contexts are included, and explains the performance gain in RG.
Also, a 69\% agreement on human-likeness demonstrates that \textsc{Caffeine}'s refinement is in line with human judgment. Refinement examples are presented in Appendix~\ref{sec:app_examples}.

\paragraph{\textsc{Caffeine} refines personas in a cost- and time-efficient manner (RQ3).}

\begin{figure}[h] 
    \centering
    \begin{minipage}{0.4\columnwidth}
        \centering
        \tiny
        \begin{tabular}{l|c|c}
        \toprule
        & Ours & ALL \\
        \midrule
        B-1 & \textbf{20.86}  & \textbf{20.86} \\
        R-1 & \textbf{20.09}  & \textbf{20.09} \\
        R-L & \textbf{15.40}  & 15.39 \\
        \bottomrule
        \end{tabular}

    \end{minipage}%
    \begin{minipage}{0.6\columnwidth}
        \centering
        \includegraphics[width=\textwidth]{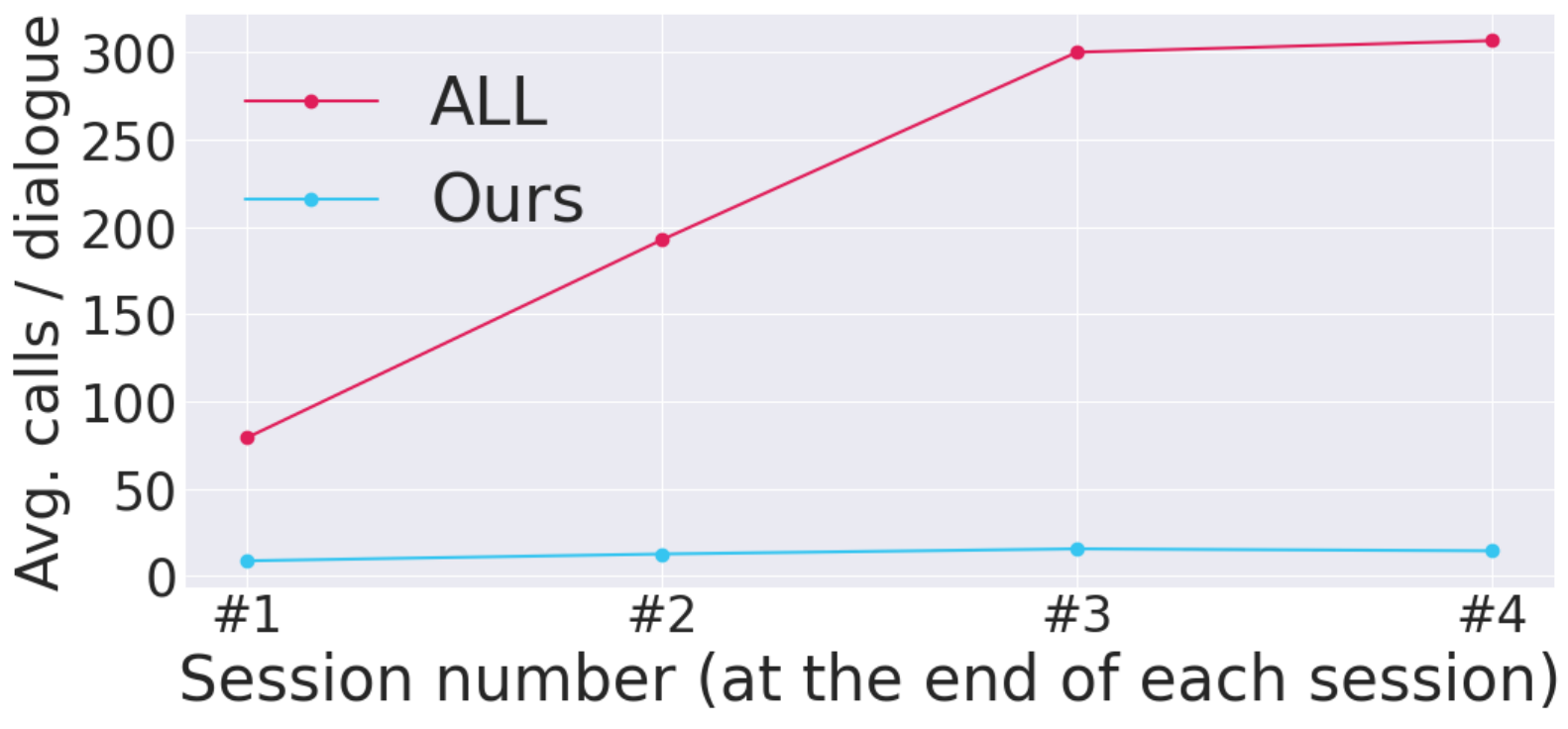}
    \end{minipage}
    \caption{Cost and time efficiency of our algorithm.}
    \label{fig:time_efficiency}
\end{figure}

 In \textsc{Caffeine}, we remove refined $(p_1, p_2)$ from $G$ after refinement. Figure~\ref{fig:time_efficiency} compares this with a setting without such removal, \ie, all $|E|$ contradictory persona pairs in $G$ are all refined (denoted as ALL). While yielding similar RG performance (Session 2-5), ours requires significantly fewer API calls per dialogue per session, especially as the sessions accumulate (9-fold $\xrightarrow{}$ 21-fold more cost- and time-efficient).

\section{Related Work}
Many studies have utilized commonsense knowledge for response generation. For instance: leveraging knowledge from a general-purpose knowledge model~\citep{zhou-etal-2022-think, wu-etal-2022-section, liu-etal-2022-think, li2023enhancing}; training commonsense generators for dialogues via human-annotated dataset~\citep{ghosal2022cicero}; formulating commonsense-linking between knowledge graphs and dialogues~\citep{gao-etal-2022-comfact}; modeling speakers' mutual beliefs before a response~\citep{zhou2022reflect}; integrating implicit information in dialogues into rationale for more effective response generation~\citep{chae2023dialogue}. While most work focuses on speaker utterances, we leverage speaker personas to address commonsense knowledge in response generation.

\section{Conclusion}
This work pioneers commonsense-based persona expansion in multi-session settings and presents a context-aware refinement framework that leverages contradictory personas to elicit a memory with richer speaker details. \coffee{} \textsc{Caffeine} improves response generation in long-term conversations and demonstrates human-like refinement of contradictory personas while being cost- and time-efficient.
\section{Limitations}
Our study has the following limitations: (1) Apart from the proposed \textsc{Caffeine}, our results can be affected by the quality of commonsense models and the knowledge graph on which they are trained. As future work, we plan to leverage LLM for persona expansion; (2) Our refinement graph stores contradictory personas that are predicted as \texttt{contradiction} with a probability higher than a pre-defined threshold by the NLI model. Our framework may miss personas that actually need a refinement due to the limitation of the NLI model; (3) While we pioneer the commonsense-based persona expansion in multi-session settings, we only consider one speaker's persona at a time in our refinement framework.
Since different people can demonstrate different personality traits and behaviors in the same commonly experienced event (\eg, discussed topic), we acknowledge there can be potential performance gain in response generation if such modeling is included; (4) In this work, we employ LLMs to generate responses based on the dialogue context and retrieved memories (\ie, both speakers' personas) in a zero-shot setting. However, since the refined personas tend to be longer and contain more information, it is possible that the LLM can not fully utilize the presented personas in its inputs as they get longer~\citep{liu2023lost}. We plan to address a better utilization of LLM's input texts for response generation in future work.
\section{Ethical Statement}
LLMs and COMET can generate sensual, harmful, biased, offensive, or violent content. Authors avoid such content from appearing in the main text, figure, and appendix.
We guarantee fair compensation for workers we hire on Amazon Mechanical Turk. We ensure an effective pay rate higher than \$18 per hour based on the estimated time required to complete the tasks.

\section{Acknowledgements}
This work was supported by Institute of Information \& Communications Technology Planning \& Evaluation (IITP) grant funded by the Korean government (MSIT)(No.2020-0-01361, Artificial Intelligence Graduate School Program (Yonsei University)) and (No.2021-0-02068, Artificial Intelligence Innovation Hub) and (No.2022-0-00077, AI Technology Development for Commonsense Extraction, Reasoning, and Inference from Heterogeneous Data). Jinyoung Yeo is a corresponding author.

\bibliography{anthology,custom}
\bibliographystyle{acl_natbib}

\appendix
\clearpage

\section{Implementation Details}
\label{sec:implementation}

\subsection{Commonsense Expansion with COMET}
\label{ssec:app_comet}
At the end of each dialogue session, we augment personas derived from the current session with new personas via COMET~\citep{hwang2021comet}, a widely used commonsense model generating rich and diverse commonsense expansions of a given statement based on cause-effect relation.
Among the 23 possible candidate relation types, following prior works on commonsense-based persona expansion~\citep{majumder2020like, kim2022dual}, we choose 9 relation types: \textsc{xAttr}, \textsc{xEffect}, \textsc{xIntent}, \textsc{xNeed}, \textsc{xReact},
\textsc{xWant}, \textsc{oEffect}, \textsc{oReact}, and \textsc{oWant} for our expansion, where the prefix ‘x’ indicates an effect or cause on that person and ‘o’ denotes others. After persona expansion via COMET, we leverage an external NLI model to initially filter out improper expansion. Specifically, when a new persona $p^n$ is generated based on an original persona $p^o$ (1 original persona yields nine 9 personas), we filter it out if the NLI model predicts the logical relationship between $p^n$ and $p^o$ is \textit{contradiction} with $\delta > 0.33$. Note that this is different from the NLI-remove baseline, as here we solely address a one-to-one relationship between a generated persona and its corresponding original persona, while the latter addresses the contradiction among all possible combinations of personas within/across the dialogue session(s). We report the statistics of this initial filtering in Table~\ref{tab:initial_filtering}.

\begin{table}[h!]
\centering
\begin{tabular}{l|cc}
\hline At the End of & Filtered (\%) & Total \\
\hline
    Session \#1 & 2830 (6.84 \%) & 41391 \\
    Session \#2 & 2715 (7.39 \%) & 36718 \\
    Session \#3 & 2935 (7.43 \%) & 39523 \\
    Session \#4 & 2971 (7.58 \%) & 39198 \\
\hline
\end{tabular}
\caption{Initial filtering of improper expansion.}
\label{tab:initial_filtering}
\end{table}

\subsection{Contriever}
In our experiments on persona-grounded response generation (RG), we adopt Contriever~\citep{izacard2021unsupervised} as the memory retriever to retrieve top-$k$ relevant personas from long-term memory based on the current conversation.
Contriever is a dense information retriever trained with unsupervised contrastive learning. Even without supervision, it has shown remarkable capabilities in information retrieval tasks, particularly in demonstrating competitiveness with BM25 in Recall at 100 (R@100) on the benchmark for zero-shot retrieval.

\subsection{Large language model}
\label{ssec:prompt}
In this work, we employ ChatGPT for the proposed \textsc{Caffeine} and response generation. ChatGPT is an LLM with 175B parameters based on InstructGPT~\citep{ouyang2022training}\footnote{https://openai.com/blog/chatgpt}. ChatGPT is trained to follow instructions given by users and return requested information in a conversational manner. 
We use LangChain\footnote{\url{https://github.com/hwchase17/langchain}} to send API calls to OpenAI API.
The prompt used in \textsc{Caffeine} and response generation are in Table~\ref{tab:prompt} and Table~\ref{tab:RG_prompt}, respectively.

\subsection{Linking Personas to their Contextual Backgrounds}
\label{ssec:dialoguefragment}
\begin{figure}[h]
    \centering
    \includegraphics[width=1\columnwidth]{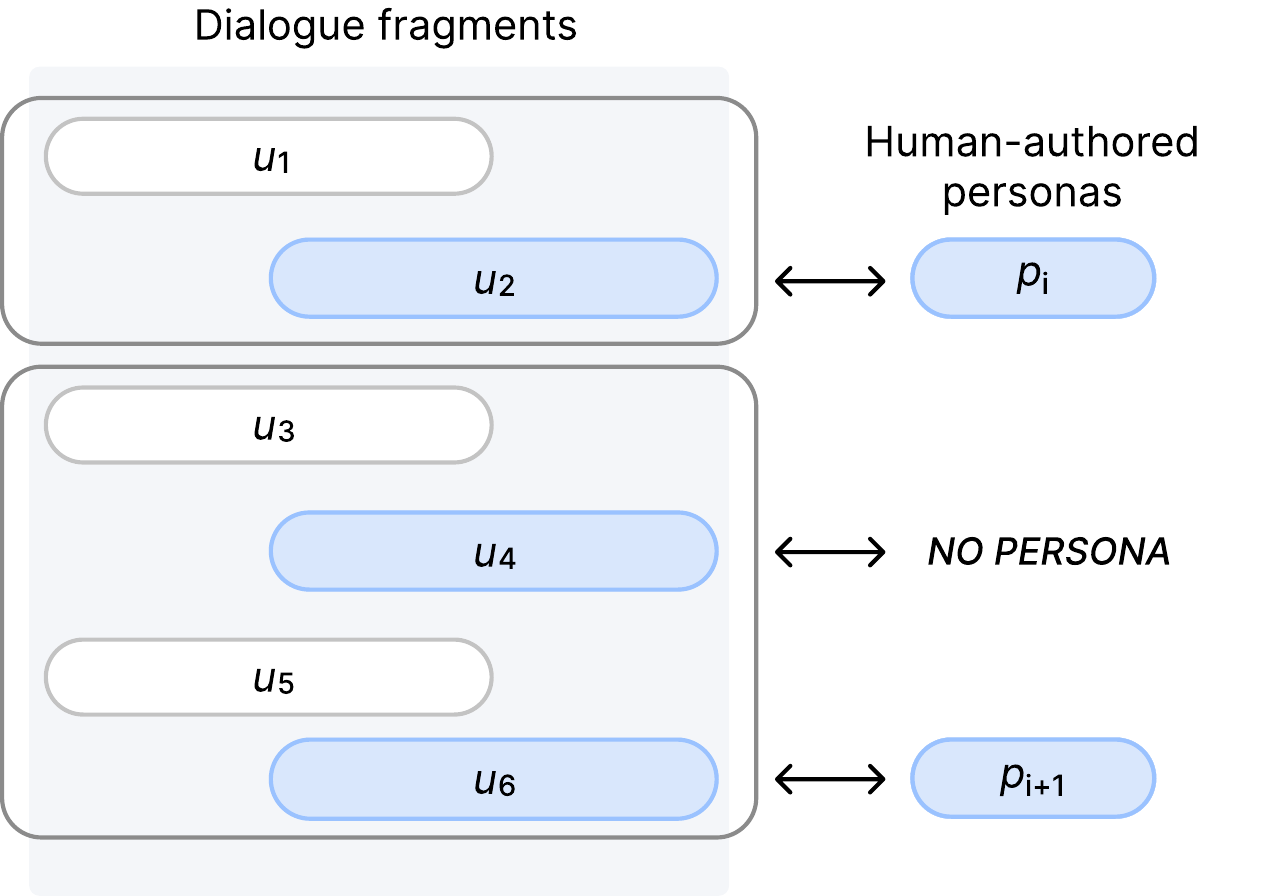}
    \caption{Demonstration of personas and their contextual backgrounds in the MSC dataset.}
    \label{fig:dialogue_fragment}
\end{figure}

In the adopted MSC dataset, human annotators summarize information in a speaker's utterance and use it to derive a persona sentence.
As demonstrated in Figure~\ref{fig:dialogue_fragment}, since not every utterance contains enough information to conclude a persona for that speaker, some utterances are not paired with a persona sentence.
In our experiment for context-aware persona refinement, we utilize contradictory personas $\mathcal{P} = (p_1, p_2)$ and their contextual backgrounds, \ie, relevant dialogue contexts $\mathcal{D} = (d_1, d_2)$ from where they are derived. $d$ consists of $w$ consecutive sentences. In practice, $w$ can differ, as we link each persona with their relevant dialogue context by separating the past conversation into dialogue fragments based on utterances that have corresponding persona sentences. For instance, the $d_i$ for persona $p_i$ will be $d_i = (u_1, u_2)$, and $d_{i+1} = (u_3, u_4, \cdots, u_6)$ for $p_{i+1}$.

\subsection{Computational Resources and API Cost}
We run Contriever and the NLI model on eight NVIDIA RTX A5000 GPUs. For ChatGPT API usage, we use \$35.52 on \textsc{Caffeine}'s refinement, and \$27.09 on response generation.

\section{Performance in Response Generation}
\label{ssec:topk}
In response generation, top-$k$ relevant persona sentences are retrieved from the long-term memory to assist response generation.
In the main text, we report the mode performance in response generation with $k = 20$, the results with $k = 12$ and  $k = 30$ are presented in Table~\ref{tab:topk_results}.

\section{Contradictory Personas in Multi-session Conversations}
\label{sec:count}
As human personalities are context-dependent, we display different personalities in different contexts and adapt to new situations. This naturally leads to personas with contradictory literal interpretations to co-exist as one's persona. Such a phenomenon does not harm human conversations. However, contradictions between personas can lead to inconsistent response generation, hindering user interest in the dialogue systems.

\begin{figure}[h]
    \centering
    \begin{subfigure}[b]{0.235\textwidth}
        \centering
        \includegraphics[width=\linewidth]{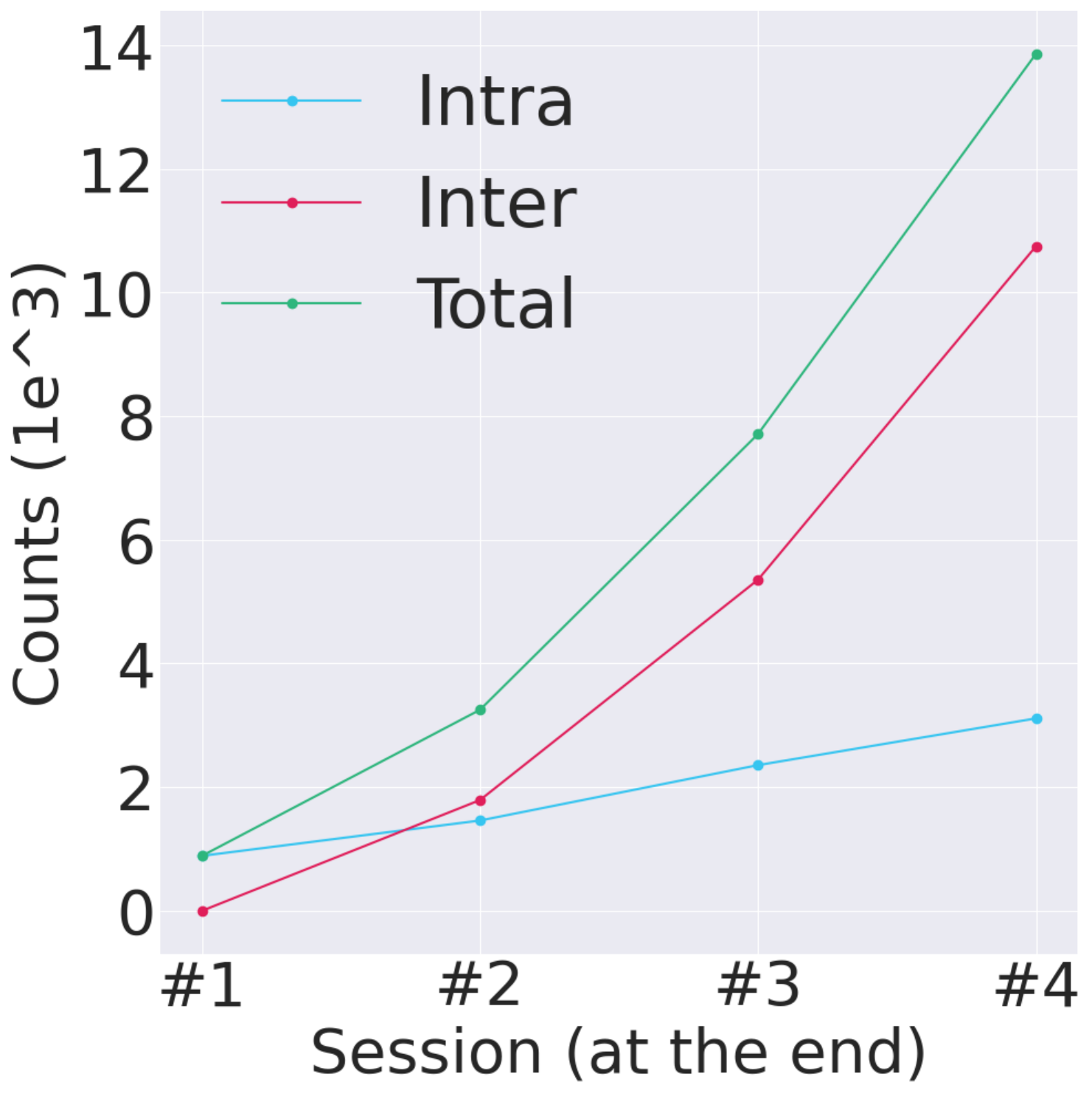} 
        \caption{GOLD}
        \label{fig:intra-session conflict}
    \end{subfigure}
    \hfill
    \begin{subfigure}[b]{0.235\textwidth}
        \centering
        \includegraphics[width=\linewidth]{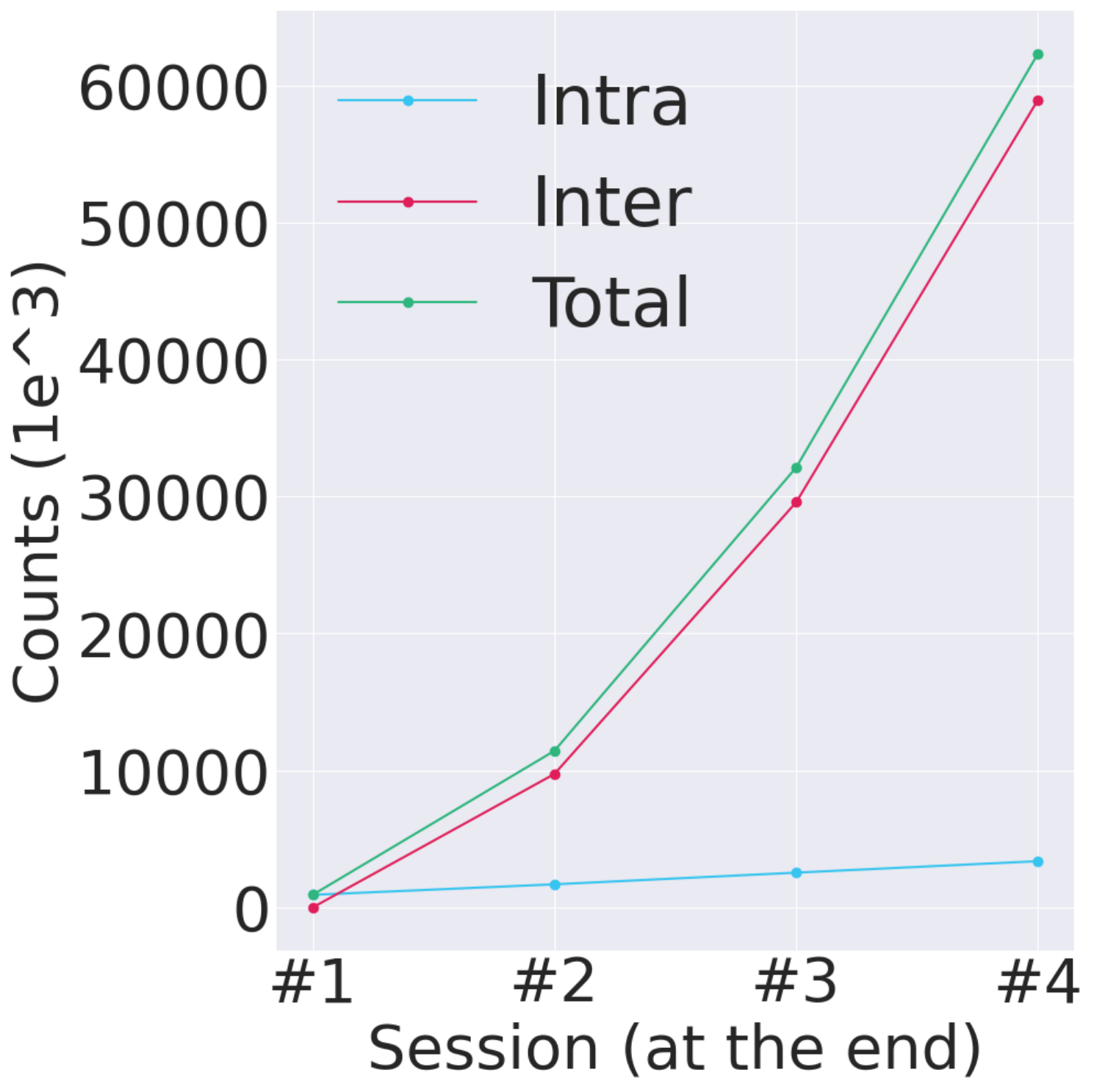}
        \caption{COMET-EXP}
        \label{fig:inter-session conflict}
    \end{subfigure}
    \caption{Contradiction among human-authored original personas (GOLD) and expanded personas (COMET-EXP). The blue, red, and green lines represent the intra-session, inter-session, and total contradictory persona pairs, respectively.}
    \label{fig:both-figures}
    
\end{figure}

In our study on the Multi-session Chat dataset, we first find that contradictory personas exist in human-authored personas (Figure~\ref{fig:both-figures} (a)). Then, we show that expanding existing human-authored personas via commonsence expansion can lead to orders-of-magnitude more contradictory personas that hinder user interest in the conversation (Figure~\ref{fig:both-figures} (b))~\citep{kim2023persona}.

Personas can contradict other personas from the same sessions (intra-session) and from the previous sessions (inter-session). When comparing COMET-EXP with the human-authored personas (GOLD), we observe that as the number of previous sessions increases, the intra-session contradiction slightly increases, whereas the inter-session contradiction skyrockets significantly. 
Although such a rising trend appears similarly in GOLD and COMET-EXP, the total count in COMET-EXP is order-of-magnitude larger.
This supports the necessity of \textsc{Caffeine}, which refines the contradictory personas in the long-term memory of dialogue models in multi-session settings.

\section{\textsc{Caffeine} vs. NLI models}
\label{ssec:nli}

\begin{figure}[h]
    \centering
    \includegraphics[width=1\columnwidth]{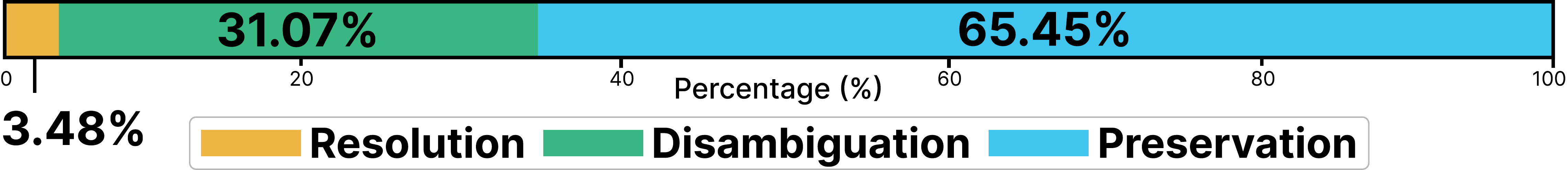}
    \caption{Proportion of selected strategies highlights the limitation of the NLI model.}
    \label{fig:refine_bar_chart}
\end{figure}

Noteworthily, Figure~\ref{fig:refine_bar_chart} shows that \textsc{Caffeine} determines that 65.45\% of contradictory personas (with $\delta \ge 0.80$) can be consistent without requiring any refinement when their contextual backgrounds are taken into account, indicating that our context-aware refinement can address the simplification of NLI models where they often solely compare the semantic representation of two statements without reasoning over their contexts. We employ two NLI models: the MNLI model (referred to as ‘NLI model’) and the DNLI model. Results with the DNLI model are presented in Table~\ref{tab:dnli_result}.

\section{Examples of Refinement and Response Generation}
\label{sec:app_examples}
\begin{figure*}
    \centering
    \includegraphics[width=1\linewidth]{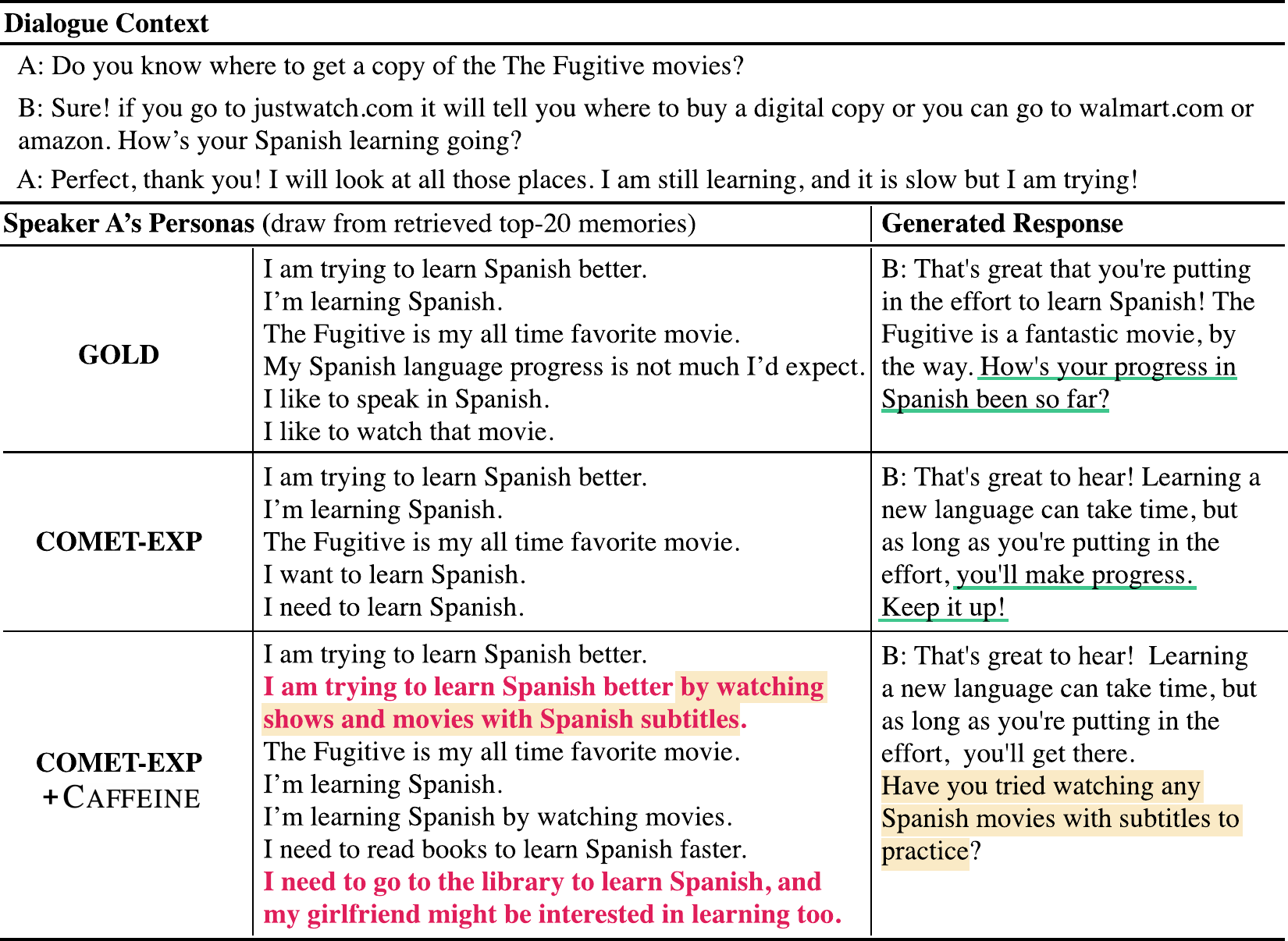}
    \caption{\textbf{Example of Response Generation \#1.} While baselines yield short personas and inconsistent or unconstructive responses (green underlines), \textsc{Caffeine} offers informative personas (color red) and leads to a response that provides constructive suggestion for Person A's Spanish learning.}
    \label{fig:appendix_rg_example}
\end{figure*}

\begin{figure*}
    \centering
    \includegraphics[width=1\linewidth]{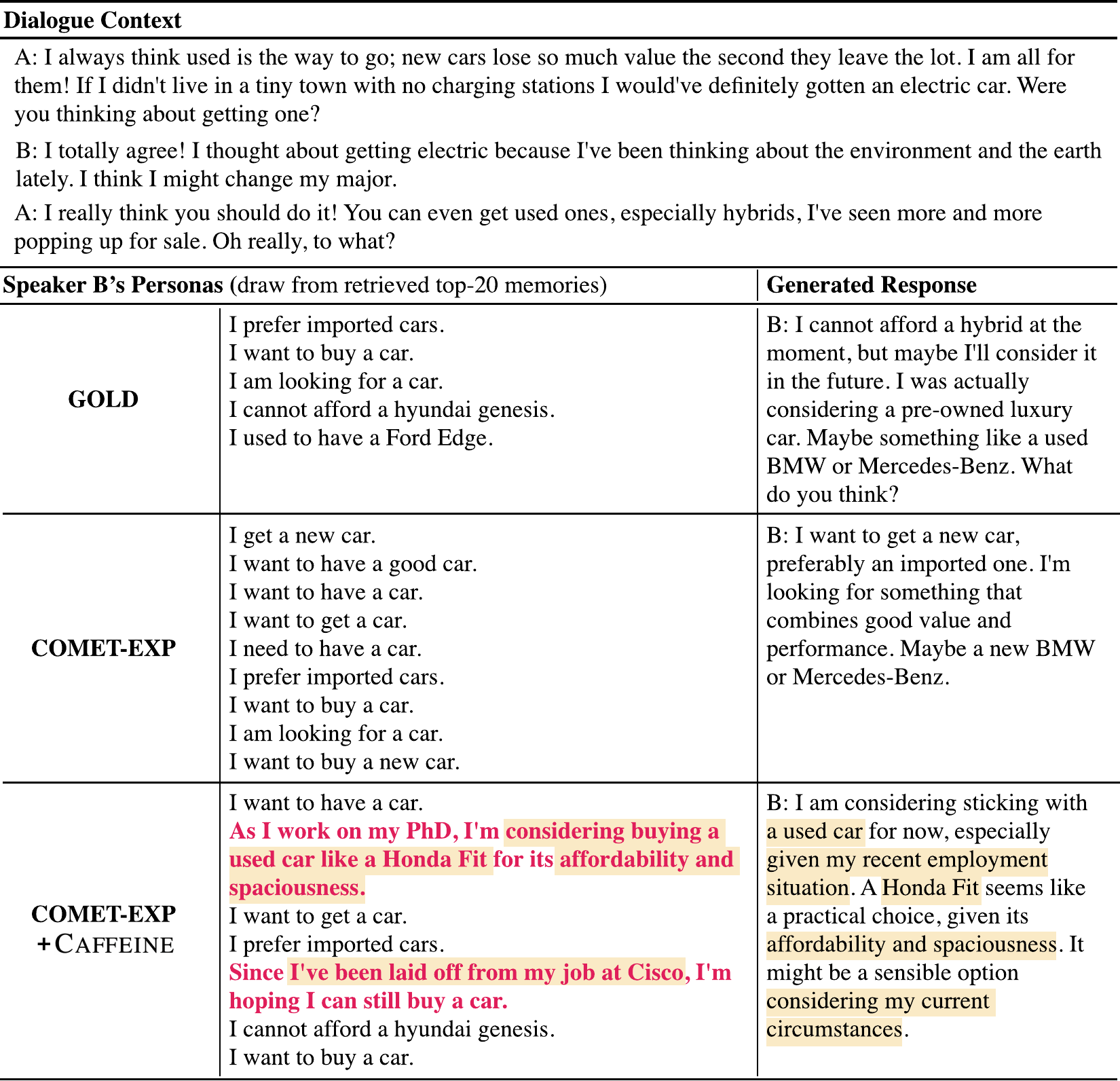}
    \caption{\textbf{Example of Response Generation \#2.} While baselines yield short personas, \textsc{Caffeine} offers informative personas (color red) and leads to a response that reflects Person B's situation and what B is looking for in a car.}
    \label{fig:appendix_rg_example_2}
\end{figure*}

We provide examples of response generation in Figure~\ref{fig:appendix_rg_example} and Figure~\ref{fig:appendix_rg_example_2}. We can observe that while baselines provide short personas and yield inconsistent or unconstructive responses (green underlines), \textsc{Caffeine} offers informative personas (color red) and leads to a response that provides constructive suggestion for Persona A's Spanish learning (Figure~\ref{fig:appendix_rg_example}) and a response that reflects Person B’s situation and what B is looking for in a car (Figure~\ref{fig:appendix_rg_example_2}).

Apart from the empirical examples demonstrated in figures in the main text, we have provided more examples for \textsc{Caffeine}'s refinement in Table~\ref{tab:app_refinement_example_1}, ~\ref{tab:app_refinement_example_3}, and~\ref{tab:app_refinement_example_4}.

\section{Details on Human Evaluation.}
\label{sec:AMT}
\subsection{Response Quality}
\begin{figure*}
    \centering
    \includegraphics[width=1\linewidth]{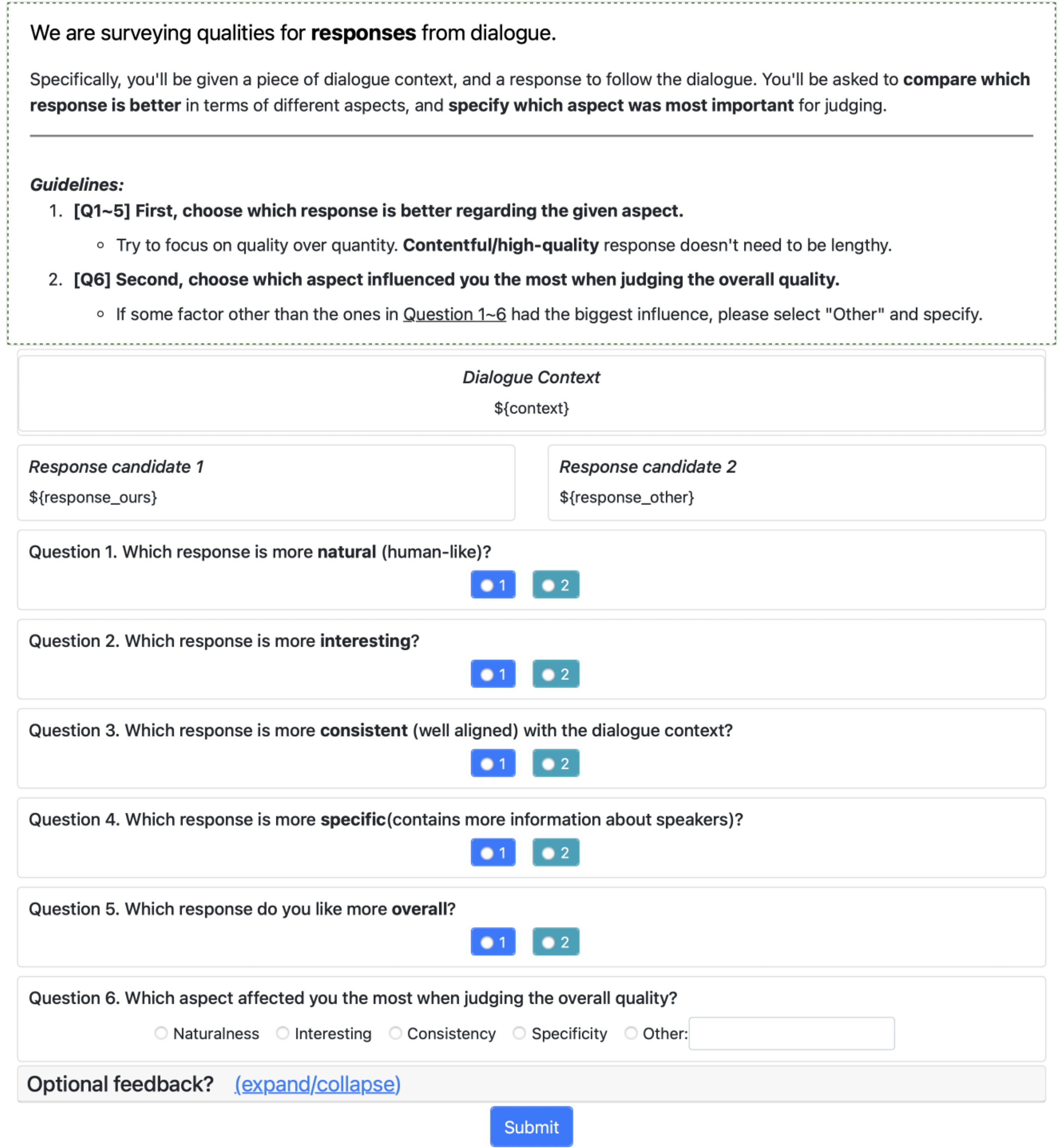}
    \caption{Interface for human evaluation on response quality.}
    \label{fig:rg_human_amt}
\end{figure*}

We outsource a human evaluation comparing the generated responses from our setting and those from the baselines via Amazon Mechanical Turk (AMT).
We show the interface for the evaluation in Figure~\ref{fig:rg_human_amt}. 
We ask the human judges to compare the responses based on the following criteria:
\begin{itemize}
    \item Naturalness: Which response is more human-like?
    \item Consistency: Which response is more consistent (aligned) with the dialogue context?
    \item Specificity: Which response contains more speaker information?
    \item Engagingness: Which response is more interesting?
 
\end{itemize}

\subsection{Refinement Quality}
\begin{figure*}
    \centering
    \includegraphics[width=1\linewidth]{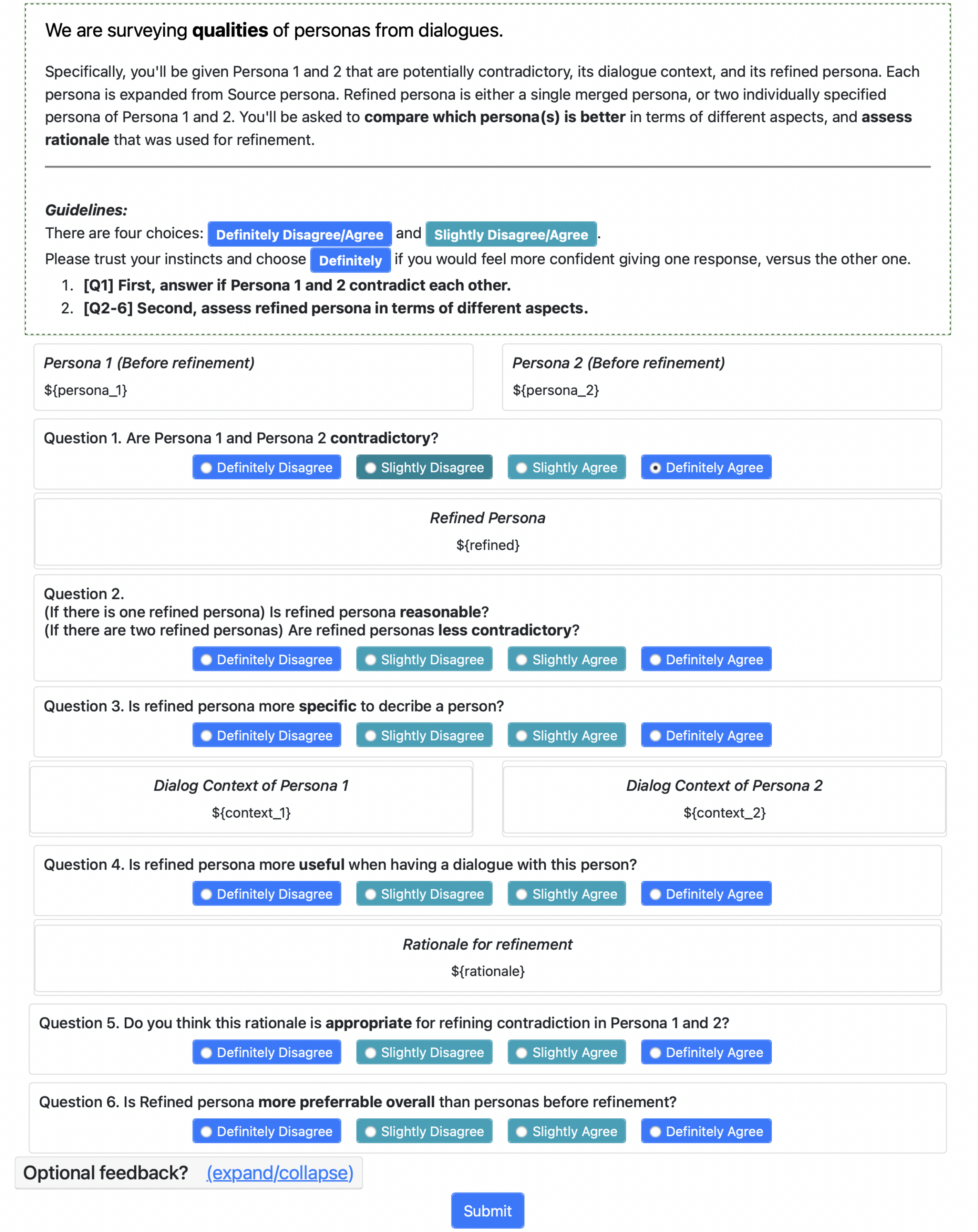}
    \caption{Interface for human evaluation on refinement quality.}
    \label{fig:refine_human_amt}
\end{figure*}

We outsource a human evaluation comparing the personas before/after \textsc{Caffeine} via Amazon Mechanical Turk (AMT).
We show the interface for this evaluation in Figure~\ref{fig:refine_human_amt}.

We ask the human judges to compare the persona before and after refinement based on the following criteria:
\begin{itemize}
    \item Consistency: Is the refined version less contradictory or more reasonable than before?
    \item Specificity: Does the refined version describe a person more specifically?
    \item Helpfulness: Can the refined version be more helpful if you are having a conversation with this person?
    \item Overall: Overall, do you prefer the refined version?
    \item Human-likeness: Is the refinement process (generated rationales) reasonable?
    
\end{itemize}
Note that before assessing \textsc{Caffeine}'s refinement, we ask workers to determine whether the personas before refinement are actually contradictory from human standpoints. The assessment of the above criteria only begins if the answer is positive. Therefore, the reported human evaluation results are based on 89 out of 100 samples we provided.

\begin{table*}[h]
\small
\centering
\begin{tabular}{lcccccccccccc}
\toprule
 &
  \multicolumn{3}{c}{\textbf{Session 2}} &
  \multicolumn{3}{c}{\textbf{Session 3}} &
  \multicolumn{3}{c}{\textbf{Session 4}} &
  \multicolumn{3}{c}{\textbf{Session 5}} \\ \cmidrule(lr){2-4} \cmidrule(lr){5-7} \cmidrule(lr){8-10} \cmidrule(l){11-13} 
\textbf{Memory} &
  B-1 &
  R-1 &
  R-L &
  B-1 &
  R-1 &
  R-L &
  B-1 &
  R-1 &
  R-L &
  B-1 &
  R-1 &
  R-L \\ 
\toprule

None & 20.75 & 19.38 & 15.16 & 20.42 & 19.53 & 15.09 & 19.88 & 19.56 & 14.98 & 19.87 & 20.16 & 15.33 \\ 
\midrule
\textbf{$k$ = 12}& & & & & & & & & \\
GOLD & \textbf{21.18} & 19.78 & \textbf{15.46} & \underline{21.26} & 20.11 & 15.43 & 20.58 & 19.97 & 15.17 & 20.38 & 20.40 & 15.42 \\ 
\hspace{0.01cm}+ NLI-remove & 20.74 & 19.83 & 15.19 & 21.05 & 20.27 & 15.51 & 21.09 & 20.35 & 15.62 & 21.22 & 20.56 & 15.78 \\ 
\cellcolor{gray!13}\hspace{0.02cm}\textbf{+ \textsc{Caffeine}} & \cellcolor{gray!13}20.91 & \cellcolor{gray!13}\underline{20.03} & \cellcolor{gray!13}\underline{15.33} & \cellcolor{gray!13}21.20 & \cellcolor{gray!13}\underline{20.52} & \cellcolor{gray!13}\textbf{15.74} & \cellcolor{gray!13}21.46 & \cellcolor{gray!13}\underline{20.77} & \cellcolor{gray!13}\textbf{15.94} & \cellcolor{gray!13}\underline{21.62} & \cellcolor{gray!13}\textbf{20.97} & \cellcolor{gray!13}\textbf{16.11} \\ 
\midrule
COMET-EXP & \underline{21.04} & 19.63 & 15.32 & 20.89 & 19.88 & 15.27 & 20.20 & 19.84 & 15.14 & 20.12 & 20.43 & 15.50 \\ 
\hspace{0.01cm}+ NLI-remove & 20.68 & 19.89 & 15.19 & 21.04 & 20.21 & 15.42 & \underline{21.49} & 20.70 & 15.81 & 21.57 & 20.73 & 15.88 \\ 
\cellcolor{gray!13}\hspace{0.02cm}\textbf{+ \textsc{Caffeine}}  & \cellcolor{gray!13}20.99 & \cellcolor{gray!13}\textbf{20.05} & \cellcolor{gray!13}15.32 & \cellcolor{gray!13}\textbf{21.41} & \cellcolor{gray!13}\textbf{20.55} & \cellcolor{gray!13}\underline{15.71} & \cellcolor{gray!13}\textbf{21.66} & \cellcolor{gray!13}\textbf{20.83} & \cellcolor{gray!13}\underline{15.93} & \cellcolor{gray!13}\textbf{21.86} & \cellcolor{gray!13}\underline{20.96} & \cellcolor{gray!13}\underline{16.07} \\ 

\midrule
\textbf{$k$ = 20}& & & & & & & & & \\
GOLD & \underline{21.19} & 19.86 & \textbf{15.50} & 21.24 & 20.16 & 15.47 & 20.57 & 19.94 & 15.16 & 20.49 & 20.53 & 15.55 \\ 
\hspace{0.01cm}+ NLI-remove & 20.81 & 19.98 & 15.26 & 21.04 & 20.28 & 15.52 & 21.33 & 20.69 & 15.91 & 21.43 & 20.75 & 15.95 \\
\cellcolor{gray!13}\hspace{0.02cm}\textbf{+ \textsc{Caffeine}} & \cellcolor{gray!13}20.93 & \cellcolor{gray!13}\textbf{20.18} & \cellcolor{gray!13}\underline{15.47} & \cellcolor{gray!13}\underline{21.41} & \cellcolor{gray!13}\underline{20.72} & \cellcolor{gray!13}\textbf{15.86} & \cellcolor{gray!13}\underline{21.67} & \cellcolor{gray!13}\underline{21.00} & \cellcolor{gray!13}\underline{16.15} & \cellcolor{gray!13}\underline{21.92} & \cellcolor{gray!13}\underline{21.23} & \cellcolor{gray!13}\underline{16.31} \\
\midrule
COMET-EXP & \textbf{21.23} & 19.82 & 15.44 & 20.95 & 19.90 & 15.38 & 20.33 & 20.02 & 15.18 & 20.00 & 20.27 & 15.37 \\ 
\hspace{0.01cm}+ NLI-remove & 20.72 & 19.96 & 15.27 & 21.12 & 20.40 & \underline{15.56} & 21.66 & 20.77 & 15.88 & 21.77 & 20.91 & 16.01 \\ 
\cellcolor{gray!13}\hspace{0.02cm}\textbf{+ \textsc{Caffeine}} & \cellcolor{gray!13}20.97 & \cellcolor{gray!13}\underline{20.06} & \cellcolor{gray!13}15.32 & \cellcolor{gray!13}\textbf{21.63} & \cellcolor{gray!13}\textbf{20.73} & \cellcolor{gray!13}\textbf{15.86} & \cellcolor{gray!13}\textbf{21.97} & \cellcolor{gray!13}\textbf{21.10} & \cellcolor{gray!13}\textbf{16.18} & \cellcolor{gray!13}\textbf{22.26} & \cellcolor{gray!13}\textbf{21.32} & \cellcolor{gray!13}\textbf{16.37} \\ 

\midrule
\textbf{$k$ = 30}& & & & & & & & & \\
GOLD & 20.88 & 19.65 & \underline{15.45} & 21.09 & 20.18 & 15.56 & 20.50 & 19.89 & 15.09 & 20.41 & 20.47 & 15.46 \\ 
\hspace{0.01cm}+ NLI-remove & 20.65 & 19.85 & 15.14 & 21.09 & 20.43 & 15.68 & 21.50 & 20.83 & 15.98 & 21.59 & 20.93 & 16.07 \\ 
\cellcolor{gray!13}\hspace{0.02cm}\textbf{+ \textsc{Caffeine}} & \cellcolor{gray!13}20.89 & \cellcolor{gray!13}\textbf{20.13} & \cellcolor{gray!13}15.43 & \cellcolor{gray!13}\underline{21.42} & \cellcolor{gray!13}\underline{20.77} & \cellcolor{gray!13}\underline{15.96} & \cellcolor{gray!13}\underline{21.73} & \cellcolor{gray!13}\underline{21.07} & \cellcolor{gray!13}\underline{16.19} & \cellcolor{gray!13}\underline{22.01} & \cellcolor{gray!13}\underline{21.29} & \cellcolor{gray!13}\underline{16.32} \\ 
\midrule
COMET-EXP & \textbf{21.40} & 19.89 & \textbf{15.52} & 21.06 & 20.10 & 15.40 & 20.38 & 20.03 & 15.27 & 20.06 & 20.50 & 15.59 \\ 
\hspace{0.01cm}+ NLI-remove & 20.60 & 19.86 & 15.20 & 21.02 & 20.33 & 15.57 & 21.35 & 20.62 & 15.83 & 21.71 & 20.97 & 16.05 \\ 
\cellcolor{gray!13}\hspace{0.02cm}\textbf{+ \textsc{Caffeine}}& \cellcolor{gray!13}\underline{20.96} & \cellcolor{gray!13}\underline{20.11} & \cellcolor{gray!13}15.37 & \cellcolor{gray!13}\textbf{21.73} & \cellcolor{gray!13}\textbf{20.85} & \cellcolor{gray!13}\textbf{15.99} & \cellcolor{gray!13}\textbf{22.17} & \cellcolor{gray!13}\textbf{21.24} & \cellcolor{gray!13}\textbf{16.27} & \cellcolor{gray!13}\textbf{22.57} & \cellcolor{gray!13}\textbf{21.58} & \cellcolor{gray!13}\textbf{16.54} \\ 
\bottomrule
\end{tabular}
\caption{Performance in response generation. \textbf{Bold} and \underline{underline} show the best and second-highest in each column.}
\label{tab:topk_results}
\end{table*}

\begin{table*}[ht]
\small
\centering
\begin{tabular}{lcccccccccccc}
\toprule
 &
  \multicolumn{3}{c}{\textbf{Session 2}} &
  \multicolumn{3}{c}{\textbf{Session 3}} &
  \multicolumn{3}{c}{\textbf{Session 4}} &
  \multicolumn{3}{c}{\textbf{Session 5}} \\ \cmidrule(lr){2-4} \cmidrule(lr){5-7} \cmidrule(lr){8-10} \cmidrule(l){11-13} 
\textbf{Settings} &
  B-1 &
  R-1 &
  R-L &
  B-1 &
  R-1 &
  R-L &
  B-1 &
  R-1 &
  R-L &
  B-1 &
  R-1 &
  R-L \\ 
\toprule

No Memory & 20.75 & 19.38 & 15.16 & 20.42 & 19.53 & 15.09 & 19.88 & 19.56 & 14.98 & 19.87 & 20.16 & 15.33 \\ 
\midrule
GOLD & \underline{21.19} & 19.86 & \textbf{15.50} & 21.24 & 20.16 & 15.47 & 20.57 & 19.94 & 15.16 & 20.49 & 20.53 & 15.55 \\ 
\hspace{0.01cm}+ DNLI-remove & 20.87 & 20.07 & 15.31 & 21.15 & 20.50 & 15.70 & 21.37 & 20.82 & 15.97 & 21.52 & 20.95 & 16.08 \\ 
\hspace{0.01cm}+ DNLI-recent & 20.92 & 20.09 & 15.36 & 21.16 & 20.58 & 15.80 & 21.36 & 20.83 & 16.01 & 21.60 & 21.08 & 16.20 \\ 
\cellcolor{gray!13}\hspace{0.01cm}\textbf{+ \textsc{Caffeine}} & \cellcolor{gray!13}20.94 & \cellcolor{gray!13}\textbf{20.15} & \cellcolor{gray!13}15.41 & \cellcolor{gray!13}21.33 & \cellcolor{gray!13}\textbf{20.69} & \cellcolor{gray!13}\underline{15.89} & \cellcolor{gray!13}\underline{21.54} & \cellcolor{gray!13}\underline{21.01} & \cellcolor{gray!13}\underline{16.17} & \cellcolor{gray!13}\underline{21.75} & \cellcolor{gray!13}21.18 & \cellcolor{gray!13}\underline{16.30} \\ \midrule
COMET-EXP & \textbf{21.23} & 19.82 & \underline{15.44} & 20.95 & 19.90 & 15.38 & 20.33 & 20.02 & 15.18 & 20.00 & 20.27 & 15.37 \\ 
\hspace{0.01cm}+ DNLI-remove & 20.81 & 20.01 & 15.26 & 21.13 & 20.46 & 15.73 & 21.53 & 20.96 & 16.12 & 21.66 & 21.06 & 16.18 \\ 
\hspace{0.01cm}+ DNLI-recent & 20.92 & \underline{20.10} & 15.41 & \underline{21.35} & \textbf{20.69} & \textbf{15.91} & 21.51 & 20.93 & 16.11 & 21.72 & \underline{21.20} & \textbf{16.34} \\ 
\cellcolor{gray!13}\hspace{0.01cm}\textbf{+ \textsc{Caffeine}} & \cellcolor{gray!13}20.89 & \cellcolor{gray!13}\underline{20.10} & \cellcolor{gray!13}15.40 & \cellcolor{gray!13}\textbf{21.37} & \cellcolor{gray!13}\underline{20.62} & \cellcolor{gray!13}15.81 & \cellcolor{gray!13}\textbf{21.82} & \cellcolor{gray!13}\textbf{21.06} & \cellcolor{gray!13}\textbf{16.19} & \cellcolor{gray!13}\textbf{22.07} & \cellcolor{gray!13}\textbf{21.21} & \cellcolor{gray!13}\underline{16.30} \\ 

\bottomrule
\end{tabular}
\caption{Performance in response generation with DNLI model. \textbf{Bold} and \underline{underline} show the best and second-highest in each column.}
\label{tab:dnli_result}
\end{table*}

\begin{table*}
    \small
    \centering
    \begin{tabular}{p{14cm}}
    \toprule
    \textbf{Prompt} \\
    \midrule
    You will be provided with two contradictory persona sentences, along with their source personas and the dialogue fragments from which these persona sentences were derived.

    Your task is to resolve the contradiction between the two persona sentences based on the dialogue fragments and the source persona of each contradictory persona. You can use these two strategies: \\
    \textbf{[Resolution]}: If the two personas are based on the same event but change over time (possibly due to a temporal difference or other events in between), adjust and aggregate them into one new persona sentence. \\
    \textbf{[Disambiguation]}: If the contradiction between them isn’t due to time changes or they are derived from unrelated events in the first place, utilize dialogue fragments to rewrite (clarify/specify) each persona. \\
    First, provide a rationale for your choice ([Resolution] or [Disambiguation]). Then, based on this rationale, generate refined persona sentence(s). 
    
    If the two personas are not contradictory, generate \textbf{[NO\_CONFLICT]}.
    \\ \midrule
    \textcolor{teal}{\textbf{Example 1:}} \\
    \textbf{\texttt{Persona 1}:} I am a programmer. \\
    Dialogue fragment of Persona 1:\\ A: As a computer programmer, I spend a lot of time writing and debugging code. It’s rewarding to see my work contribute to the development of functional and efficient software. \\
    \textit{Source Persona:} I am a programmer. \\
    
    \textbf{\texttt{Persona 2:}} I get fire. \\
    Dialogue fragment of Persona 2:\\ A: I got fired. It was quite unexpected, and I'm still processing everything. \\B: I’m really sorry to hear that. Do you want to talk about what happened? \\
    \textit{Source Persona:} I don't have a job right now. \\
    
    \textbf{\texttt{Rationale:}} There is a temporal connection between the two personas. Persona 1 is about being a programmer, whereas Persona 2 is about having been fired. Both personas can exist over time with Persona 2 occurring after Persona 1. \\
    \textbf{[Resolution]:} I am a programmer who has recently been fired. \\\midrule
    
    \textcolor{teal}{\textbf{Example 2:}} \\
    \textbf{\textit{\texttt{Persona 1}}:} I feel happy. \\
    Dialogue fragment of Persona 1: \\A: My favorite baseball team won this season! I'm so excited! \\B: Wow, really? That's great news! \\
    \textit{Source Persona:} My team won the league. \\
    
    \textbf{\texttt{Persona 2:}} I have been feeling quite stressed. \\
Dialogue fragment of Persona 2: \\A: Work has been overwhelming with tight deadlines and high expectations. It’s been hard to find time for relaxation and self-care. \\B: With stress piling up, it’s important to find ways to unwind and destress. Maybe I’ll try meditation or yoga to help manage my anxiety. \\
    \textit{Source Persona:} I work very hard. \\
    
    \textbf{\texttt{Rationale:}} The two personas do not reflect changes over time but rather different emotional states in response to separate circumstances; one, a moment of happiness due to a favorite team winning, and the other, underlying stress caused by work pressures. \\
    \textbf{[Disambiguation]:} \\
    - \textbf{{\texttt{Persona 1:}}} I feel happy when my favorite baseball team wins. \\
    - \textbf{\texttt{Persona 2:}} I am a person dealing with work-related stress and looking for ways to manage anxiety.\\ \midrule
    \textbf{\textcolor{teal}{\textbf{Example 3:}}} \\
    \textbf{\textit{\texttt{Persona 1}}:} I am a vegetarian. \\
    Dialogue fragment of Persona 1: \\
    A: I’ve been a vegetarian for about 5 years now. It started for health reasons, but now it's also about sustainability and animal welfare. \\
    \textit{Source Persona:} I am a vegetarian. \\
    
    \textbf{\texttt{Persona 2:}} I enjoy reading fiction books. \\
    Dialogue fragments of Persona 2: \\A: One of my favorite pastimes is reading fiction. I love getting lost in different worlds and lives through the pages of a good novel. \\
    \textit{Source Persona:} I enjoy reading fiction books. \\
    
    \textbf{\texttt{Rationale:}} The two persona sentences do not contradict each other as they pertain to different aspects of the speaker’s identity. One persona is about dietary preference (being a vegetarian), and the other is about a hobby or interest (enjoying reading fiction books). There is no inherent conflict between being a vegetarian and enjoying reading fiction, so the two persona sentences can coexist without the need for resolution or disambiguation. \\
    \textbf{[NO\_CONFLICT]} \\ \midrule
    \textcolor{teal}{\textbf{Example 4:}} ... \\ \bottomrule
    \end{tabular}
    \caption{The prompt for \textbf{\textsc{Caffeine}} (Five-shot setting, Examples 4 and 5 are omitted in this table). The ``preservation'' strategy is represented as \textbf{[NO\_CONFLICT]} in our prompt.}
    \label{tab:prompt}
\end{table*}

\begin{table*}
    \small
    \centering
    \begin{tabular}{p{14cm}}
    \toprule
    \textbf{Prompt} \\
    \midrule
    You will be generating the next turn of a given dialogue context between Speaker A and Speaker B. Alongside the dialogue context, you'll be given persona statements about both speakers. Your response should be 1-2 sentences, utilizing the persona statements as guidance to create an appropriate reply. Generate appropriate answers using given persona statements as memory.\\What is the most appropriate next utterance (3 sentences max)?
    \\
    \\
    \textbf{\texttt{Persona Statements of A}:} \textcolor{teal}{\{A's personas within the top-$k$ retrieved personas from long-term memory\}} \\
    \textbf{\texttt{Persona Statements of B}:} \textcolor{teal}{\{B's personas within the top-$k$ retrieved personas from long-term memory\}} \\
    \textbf{\texttt{Dialogue}:}  \textcolor{teal}{\{dialogue context\}} \\
    \textbf{\texttt{Response}:} \textcolor{magenta}{\{generation result\}}
    \\ \bottomrule
    \end{tabular}
    \caption{The prompt for response generation (zero-shot).}
    \label{tab:RG_prompt}
\end{table*}

\begin{table*}[t] 
\centering
\begin{tabular}{|p{15cm} l|}
\hline
\textbf{Example 1 - [Resolution]}& \\
\hline
\textbf{Before refinement:} & \\
\texttt{B's Persona 1:} I feel happy.  &\\
\texttt{B's Persona 2:} I feel sad. &\\ &\\
Contextual Backgrounds (relevant dialogue contexts):& \\
\textcolor{gray}{In this example, both personas are derived from the same dialogue context.} & \\
 A: It's really bad from the first of the year until tax deadline day but it keeps me pretty busy the rest of the year as well. I will need to think it over. It does get lonely sometimes.&\\
 B: I can only imagine. That was \textcolor{magenta}{another thing about being a vet that was hard}. People expected you to be on call all the time, so I \textcolor{magenta}{didn't have a chance for any sort of social life}. That's gotten a lot better since I quit. \textcolor{magenta}{I've found a good group of friends to hang out with} at least a few times a week at a local cafe.&\\
 &\\
 \textbf{\textsc{Caffeine}'s output:} &\\
 \textbf{\texttt{Explanation:}} The two persona sentences reflect contradictory emotional states. Persona 1 expresses happiness due to having a good group of friends to hang out with, whereas Persona 2 expresses sadness and loneliness from lacking a social life. Both personas are based on the same events and experiences, indicating a change in emotional state over time. & \\
\textbf{\texttt{After refinement:}} I \textcolor{magenta}{used to feel sad and lonely when I was a vet}, but \textcolor{magenta}{now I feel happy because I have a good group of friends to hang out with} at a cafe every week.& \\
\hline
\hline
\textbf{Example 2 - [Disambiguation]}& \\
\hline
\textbf{Before refinement:} & \\
\texttt{B's Persona 1:} I want to play with the dogs.  &\\
\texttt{B's Persona 2:} I want to eat them. &\\ &\\
Contextual Backgrounds (relevant dialogue contexts):& \\
 A: Yeah definitely, that's great to hear you're enjoying your new vegan diet. Oh, cute! I love labs. Yes, my puppies keep me busy as well. The training phase is definitely a lot of work. \textcolor{magenta}{How old are your labs now}? &\\
 B: \textcolor{magenta}{Almost 4. Still act like puppies running around like crazy}. I think I \textcolor{magenta}{wouldn't get any exercise if it weren't for them} though, so that is a good thing for me.
 &\\
 - - - - -
 & \\
 A: That sounds delicious!!! I agree, a recipe exchange would be great. True, vegan cheese does help with that. \textcolor{magenta}{Have you found any favorite brands for vegan cheese}? I found a couple that I like, and some that I didn't enjoy as much. & \\
 B: \textcolor{magenta}{I have tried a couple, none that I've enjoyed so far}. the next on my list to try is cashew cheese, which I have heard is superior, \textcolor{magenta}{so I am excited to try it}. I hear it makes great nachos, and I miss nachos so much! & \\ &\\
 \textbf{\textsc{Caffeine}'s output:} &\\
 \textbf{\texttt{Explanation:}} The two personas do not reflect changes over time but rather different topics of discussion. Persona 1 is about the speaker's dogs and their age, while Persona 1 is about discussing vegan cheese and favorite brands.  & \\
\textbf{\texttt{After refinement:}} & \\ 
\texttt{B's Persona 1:} My \textcolor{magenta}{4-year-old labs (labrador) keep me active} and it's a good thing for me.& \\ 
\texttt{B's Persona 2:} I \textcolor{magenta}{enjoy trying different brands of vegan cheese}, although \textcolor{magenta}{I haven't found a favorite yet}.& \\

\hline
\end{tabular}

\caption{More examples of persona refinement done by \textbf{\textsc{Caffeine}}. \textcolor{magenta}{Color magenta} highlights the contextual cues from the relevant dialogue contexts.}
\label{tab:app_refinement_example_1}

\end{table*}

\begin{table*}[t] 
\centering
\begin{tabular}{|p{15cm} l|}
\hline
\textbf{Example 1 - [Resolution]}& \\
\hline
\textbf{Before refinement:} & \\
\texttt{A's Persona 1:} I feel excited. &\\
\texttt{A's Persona 2:} I get tired. &\\ &\\
Contextual Backgrounds (relevant dialogue contexts):& \\
\textcolor{gray}{In this example, both personas are derived from the same dialogue context.} & \\
B: I am sure you feel so small and at peace up there. When will your mission make its trip? I imagine \textcolor{magenta}{training for space is long and strenuous.}  & \\
A: It absolutely is! The trip is scheduled for around this time next year. \textcolor{magenta}{I'm going to have to get on a pretty strict diet and exercise routine} to make sure that I am fit enough for the trip. & \\
 &\\
 \textbf{\textsc{Caffeine}'s output:} &\\
 \textbf{\texttt{Explanation:}} The two persona sentences contradict each other in terms of the emotional state of the speaker. One persona expresses excitement, while the other expresses tiredness. However, both personas are based on the same event or circumstance of training for a space mission. Thus, there is a temporal aspect to this contradiction. & \\
\textbf{\texttt{After refinement:}} 
I am currently feeling tired due to the \textcolor{magenta}{strenuous training for my upcoming space mission}, but I am also \textcolor{magenta}{excited about the opportunity}.& \\
\hline
\hline
\textbf{Example 2 - [Disambiguation]}& \\
\hline
\textbf{Before refinement:} & \\
\texttt{A's Persona 1:} I feel relaxed. &\\
\texttt{A's Persona 2:} I feel tired. &\\ &\\
Contextual Backgrounds (relevant dialogue contexts):& \\

B: That sounds like so much fun. I have never been fishing in my life. Is it difficult to learn? &\\
A: There's a bit of a knack to the rod, but once you get that down, it's pretty simple. It's a lot of just sitting around and waiting. \textcolor{magenta}{It's very peaceful and relaxing.} &\\
- - - - - &\\
B: We are always on the lookout for a new restaurant. Have you tried a great place recently? &\\
A: \textcolor{magenta}{ I have been so busy with work}, that I haven't! I was hoping you might have a recommendation. We definitely need to go somewhere with great crab legs, or else a nice sushi joint. Somewhere with delicious cocktails as well!&\\

 &\\
 \textbf{\textsc{Caffeine}'s output:} &\\
 \textbf{\texttt{Explanation:}} The two persona sentences do not reflect changes over time but rather different emotional states and interests. Persona 1 expresses feeling happy, while Persona 2 expresses being bored. The contradiction lies in the difference in emotional experiences and preferences. & \\
\textbf{\texttt{After refinement:}} & \\ 
\texttt{A's Persona 1:} I feel relaxed \textcolor{magenta}{when I go fishing.}& \\ 
\texttt{A's Persona 2:} I feel tired because \textcolor{magenta}{I spend a lot of time at work.}& \\

\hline
\end{tabular}

\caption{More examples of persona refinement done by \textbf{\textsc{Caffeine}}. \textcolor{magenta}{Color magenta} highlights the contextual cues from the relevant dialogue contexts.}
\label{tab:app_refinement_example_3}

\end{table*}

\begin{table*}[t] 
\centering
\begin{tabular}{|p{15cm} l|}
\hline
\textbf{Example 1 - [Preservation]}& \\
\hline
\textbf{Before refinement:} & \\
\texttt{B's Persona 1:} I love punk music. &\\
\texttt{B's Persona 2:} I enjoy romantic comedies and would like to watch some cop shows. &\\ &\\
Contextual Backgrounds (relevant dialogue contexts):& \\
A: I am great. Residing in Arizona, skydiving is awesome, Barnes \& Noble is my favorite, you? &\\
B: I like movies over books, \textcolor{magenta}{love punk music} and just love to cruise around!& \\
- - - - - &\\
A: I read the Tori Spelling novel you told me about! &\\
B: Awesome! What did you think of it? &\\
A: I thought it was really good. Thanks for recommending it. You also said you watch a lot of movies. Can you recommend a good movie to me too? &\\
B: Well, \textcolor{magenta}{I've been watching a lot of romantic comedies lately.} We should watch a couple the next time we have a movie night! I also wanted to ask, \textcolor{magenta}{are there any cop shows you'd recommend? I'd like to try watching them too.}&\\
&\\
 \textbf{\textsc{Caffeine}'s output:} &\\
 \textbf{\texttt{Explanation:}} The two persona sentences do not contradict each other as they pertain to different areas of interest. Persona 1 expresses a love for punk music, while Persona 2 expresses a liking for romantic comedies and an interest in watching cop shows. There is no inherent conflict between loving punk music and enjoying romantic comedies or cop shows, so the two persona sentences can coexist without the need for resolution or disambiguation.& \\
\hline
\hline
\textbf{Example 2 - [Preservation]}& \\
\hline
\textbf{Before refinement:} & \\
\texttt{B's Persona 1:} I got a new car. &\\
\texttt{B's Persona 2:} I like to drink water. &\\ &\\
Contextual Backgrounds (relevant dialogue contexts):& \\
A: \textcolor{magenta}{How was the drive up to San Diego?} Did you bring your kids or are you flying solo? I hear the beaches are beautiful there. &\\
B: The drive was long, but the scenery was gorgeous! Took a lot longer to make the trip up than I expected because \textcolor{magenta}{with the kids} I had to make more stops to stretch our legs, feed them, and give them a little time to burn off some energy. I think I always took for granted the luxury of traveling sans kids in the past, but honestly I don't mind the company! It was the first time traveling with my kids \textcolor{magenta}{in my new car.} How's it going for you?&\\
- - - - - &\\
A: I have not been to San Diego, but I have always wanted to go. You will have to tell me all about it. &\\
B: I will. I will be on the lookout for the best iced tea! Haha. 
A: You know it is one of my favorites to drink. Besides tea I only really drink water  &\\
B: \textcolor{magenta}{I love water, too.} I wish I could get my kids to drink more water.&\\
 &\\
 \textbf{\textsc{Caffeine}'s output:} &\\
 \textbf{\texttt{Explanation:}} The two personas do not contradict each other as they pertain to different aspects of the speaker's life. Persona 1 is about getting a new car and taking a trip to San Diego, while Persona 2 is about the speaker's preference for drinking water. The two persona sentences can coexist without the need for resolution or disambiguation. & \\
\hline
\end{tabular}
\caption{Examples of ``preservation'' addressing the sub-optimal performance of NLI models that solely rely on the persona sentences without contextual backgrounds. \textcolor{magenta}{Color magenta} highlights the contextual cues.}
\label{tab:app_refinement_example_4}

\end{table*}

\end{document}